\documentclass[letterpaper]{article} 
\usepackage{aaai24}  
\usepackage{times}  
\usepackage{helvet}  
\usepackage{courier}  
\usepackage[hyphens]{url}  
\usepackage{graphicx} 
\usepackage{natbib}  
\usepackage{caption} 
\usepackage{algorithm}

\usepackage{newfloat}
\usepackage{listings}

\usepackage{graphicx}
\usepackage{subfig}
\usepackage{booktabs} 
\usepackage{pgfplots}

\usepackage{amsmath}
\usepackage{algpseudocode}

\newtheorem{lemma}{Lemma}
\newtheorem{lemma_app}{Lemma}

\newtheorem{remark}{Remark}
\newtheorem{definition}{Definition}

\newtheorem{proof}{Proof}

\usepackage{pifont}
\newcommand{\cmark}{\ding{51}}%
\newcommand{\xmark}{\ding{55}}%

\renewcommand{\div}{\nabla\cdot\,}

\newcommand{\std}[1]{\pm #1}
\usepackage{multirow}
\newcommand{\bfA}{{\bf A}}

\newcommand{\bfD}{{\bf D}}

\newcommand{\bfI}{{\bf I}}

\newcommand{\bfK}{{\bf K}}
\newcommand{\bfL}{{\bf L}}

\newcommand{\bfQ}{{\bf Q}}
\newcommand{\bfR}{{\bf R}}

\newcommand{\bfT}{{\bf T}}
\newcommand{\bfU}{{\bf U}}
\newcommand{\bfV}{{\bf V}}

\newcommand{\bfX}{{\bf X}}
\newcommand{\bfY}{{\bf Y}}
\newcommand{\bfZ}{{\bf Z}}

\newcommand{\bfv}{{\bf v}}

\newcommand{\bftheta}{{\boldsymbol \theta}}

\newcommand{\first}[1]{\textbf{\textcolor{black}{#1}}}
\newcommand{\second}[1]{\text{\textcolor{black}{#1}}}
\newcommand{\third}[1]{\text{\textcolor{black}{#1}}}

\usepackage{tikz}
\usepackage{amsfonts}
\usepackage{graphicx}
\usepackage{caption}

\DeclareCaptionStyle{ruled}{labelfont=normalfont,labelsep=colon,strut=off} 
\floatstyle{ruled}
\newfloat{listing}{tb}{lst}{}
\floatname{listing}{Listing}
\title{Feature Transportation Improves  Graph Neural Networks}
\author{
    Moshe Eliasof\textsuperscript{\rm 1,2}\thanks{Work done while being a Ph.D. student at Ben-Gurion University of the Negev.}, 
    Eldad Haber\textsuperscript{\rm{3}}, Eran Treister\textsuperscript{\rm{2}}
}
\affiliations{
    \textsuperscript{\rm 1}
    Department of Applied Mathematics and Theoretical Physics, University of Cambridge, United Kingdom\\
    \textsuperscript{\rm 2}
    Department of Computer Science, Ben-Gurion University of  the Negev, Israel\\
        \textsuperscript{\rm 3}
    Department of Earth, Ocean, and Atmospheric Sciences, University of British Columbia, Canada\\
    me532@cam.ac.uk, 
    ehaber@eoas.ubc.ca,
    erant@cs.bgu.ac.il 

}

\usepackage{bibentry}

\begin{document}

\maketitle

\begin{abstract}
Graph neural networks (GNNs) have shown remarkable success in learning representations for graph-structured data. However, GNNs still face challenges in modeling complex phenomena that involve feature transportation. In this paper, we propose a novel GNN architecture inspired by Advection-Diffusion-Reaction systems, called ADR-GNN.
Advection models feature transportation, while diffusion captures the local smoothing of features, and reaction represents the non-linear transformation between feature channels. We provide an analysis of the qualitative behavior of ADR-GNN, that shows the benefit of combining advection, diffusion, and reaction.
To demonstrate its efficacy, we evaluate ADR-GNN on real-world node classification and spatio-temporal datasets, and show that it improves or offers competitive performance compared to state-of-the-art networks.
\end{abstract}

\section{Introduction}
\label{sec:intro}

Recently, GNNs have been linked to ordinary and partial differential equations (ODEs and PDEs) in a series of works \cite{ zhuang2020ordinary,chamberlain2021grand, eliasof2021pde, rusch2022graph, wang2022acmp, di2022graphGRAFF, gravina2022anti}. 
These works propose to view GNN layers as the time discretization of ODEs and PDEs, and as such they offer both theoretical and practical advantages. First, ODE and PDE based models allow to reason about the behavior of existing GNNs. For instance, as suggested in \cite{chamberlain2021grand}, it is possible to view GCN \cite{kipf2016semi} and GAT \cite{velickovic2018graph} as discretizations of the non-linear heat equation. This observation helps to analyze and understand the oversmoothing phenomenon in GNNs \cite{nt2019revisiting,oono2020graph,cai2020note}. Second, ODE and PDE based GNNs pave the path to the construction and design of GNNs that satisfy desired properties, such as energy-preservation \cite{eliasof2021pde, rusch2022graph}, attraction and repulsion forces modeling \cite{wang2022acmp, di2022graphGRAFF}, anti-symmetry \cite{gravina2022anti}, as well as reaction-diffusion systems \cite{choi2022gread}.
Nonetheless, the aforementioned architectures still rely on controlled diffusion or wave propagation, as well as non-linear pointwise convolutions. Therefore, as discussed in \cite{rusch2023survey}, while there are methods that can alleviate oversmoothing, they may lack expressiveness. 
We now provide a simple example, known as the graph node feature transportation task \cite{leveque}, where diffusion, wave propagation, and reaction networks may fail. In this task, the goal is to gather the node information (i.e., features) from several nodes to a single node. Clearly, no diffusion process can express or model such a phenomenon, because diffusion spreads and smooths, rather than transports information \cite{EvansPDE, ascher2008numerical}. Likewise, a wave-propagation approach cannot express such a phenomenon, because it lacks directionality \cite{ascher2008numerical}, which is required for this task.
An instance of this problem is illustrated in Figure \ref{fig:transportExample}, where we show the source and target node features, and the learned advection weights that can achieve the desired target. Later, in Figure \ref{fig:syntheticExample}, we show that popular operators such as diffusion or reaction cannot model the transition from the source to the target node features, while advection can.
Furthermore, the concept of advection appears in many real-world problems and data, such as traffic-flow and-control \cite{BetJ:01}, quantity transportation in computational biology \cite{uys2009coupling}, and rainfall forecasting \cite{seed2003dynamic}.
Motivated by the previously discussed observations and examples, we propose, in addition to learning and combining diffusion and reaction terms, to develop a learnable, neural \emph{advection} term, {also known as a \emph{transportation} term \cite{EvansPDE, ascher2008numerical, leveque}, that is suited to model feature transportation from the data in a task driven fashion.
The resulting architecture, called \emph{ADR-GNN}, can therefore express various phenomena, from advection, diffusion, to pointwise reactions, as well as their compositions.

\begin{figure}   
\centering
        \subfloat[Advection]{
\includegraphics[width=0.13\textwidth]{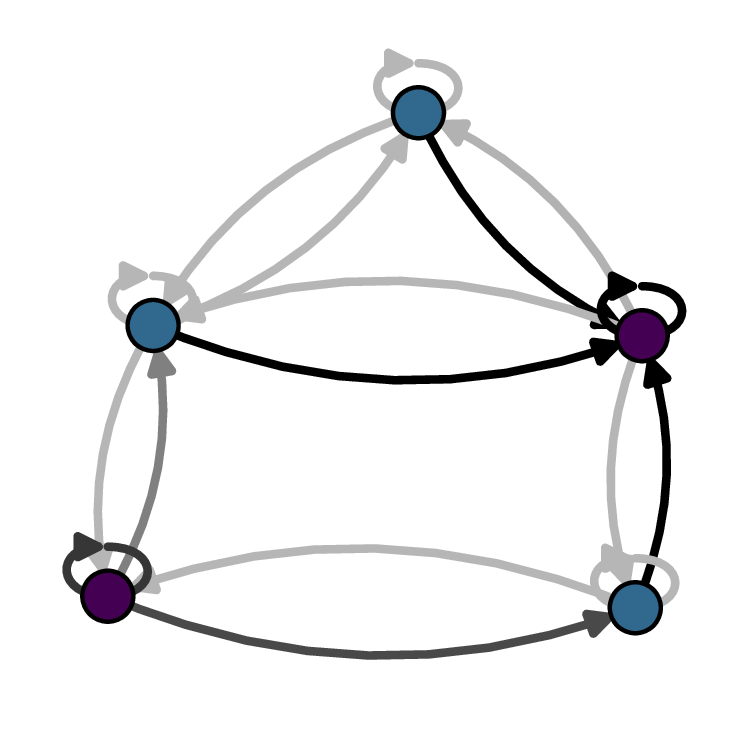}}
    \subfloat[Source]{    
\includegraphics[width=0.13\textwidth]{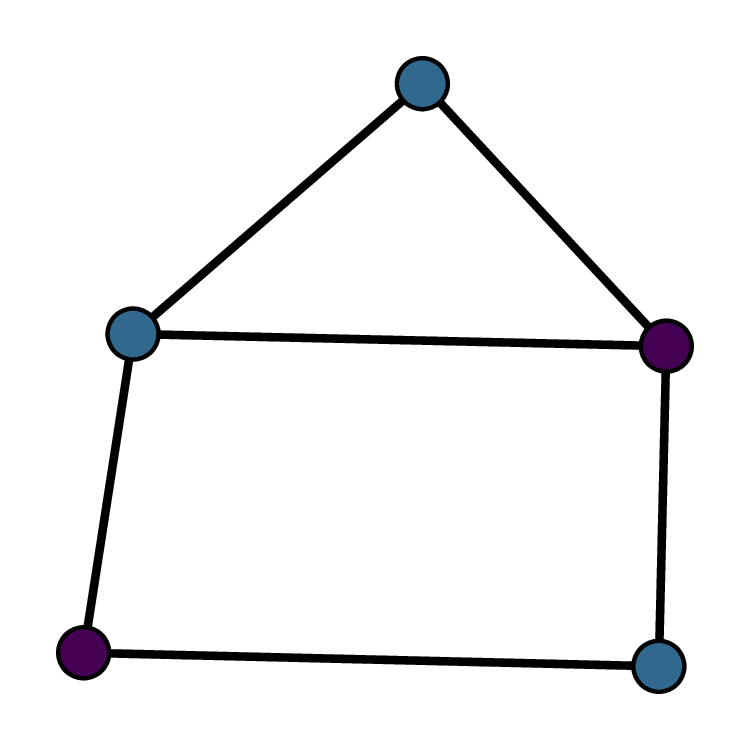}}
    \subfloat[Target]{  
\includegraphics[width=0.13\textwidth]{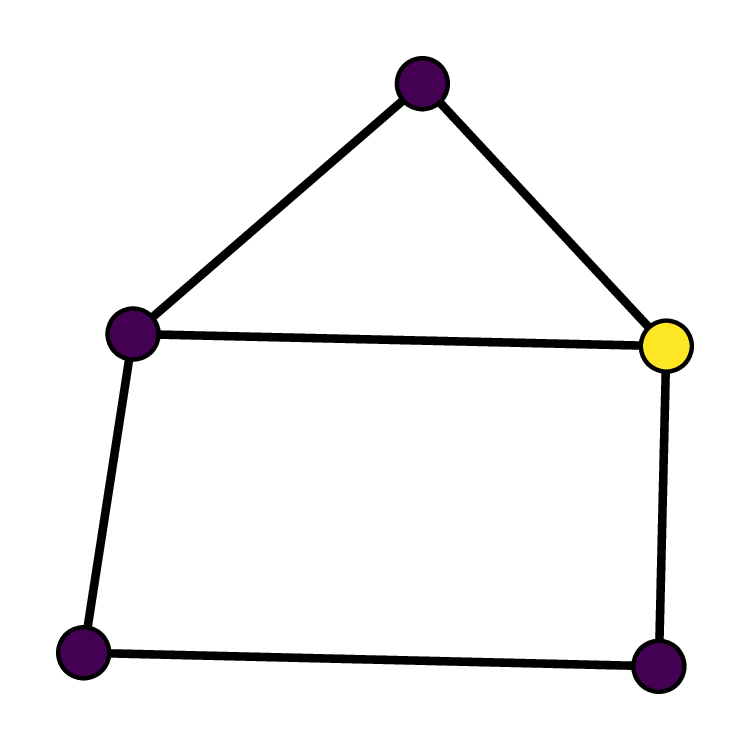}}
\includegraphics[width=0.05\textwidth,trim={0cm 0cm 7.5cm 0},clip]{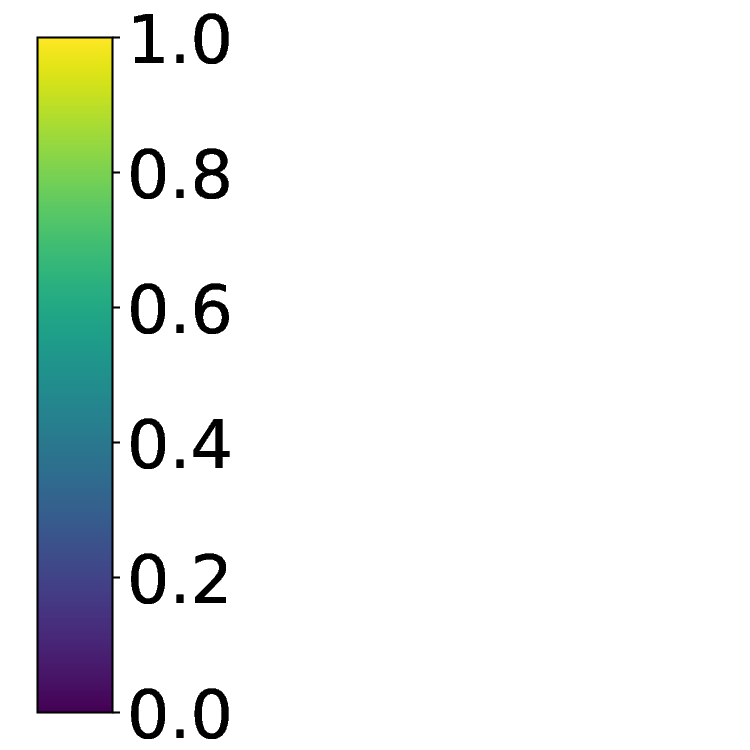}
\caption{An example of node feature transportation on a graph. Applying the advection weights in (a) to the source (b), yields the target (c). Darker edge colors in (a) indicate greater advection weights.}
\label{fig:transportExample}
\end{figure}

\textbf{Contributions.} 
The contributions of this paper are three-fold. (1) We develop a novel graph neural advection operator that is mass preserving, stable, and consistent with continuous advection PDEs. This operator enables the modeling of phenomena that involve feature transportation on graphs by learning the direction of the transportation. (2) We propose ADR-GNN, a GNN based on learnable advection-diffusion-reaction (ADR) systems, that can express a wide range of phenomena, including learned directional information flow, diffusion, and pointwise reactions. (3) We demonstrate the efficacy of ADR-GNN on node classification, and spatio-temporal forecasting datasets, achieving improved or competitive results compared to state-of-the-art models. 

\section{Related Work}
\label{sec:related}
\textbf{Advection-Diffusion-Reaction.} An Advection Diffusion Reaction system is a mathematical model that describes the simultaneous existence of three  processes: (1) the advection (transport) of information in a medium, (2) the diffusion (smoothing) of information within that medium, and (3) pointwise (self) reactions. These systems are used to study and model a wide range of physical, chemical, and biological phenomena. For example, ADR systems can be utilized to track and estimate the location of fish swarms \cite{adam2004use}, modeling ecological trends \cite{cosner2014reaction}, and the modeling of turbulent flames in supernovae \cite{khokhlov1995propagation}.  However, the aforementioned works rely on a low-dimensional, hand-crafted, non-neural ADR system to be determined, typically by trial and error, often requiring a domain expert. 
In contrast, in this paper we propose to learn the ADR system for various graph types and tasks.

\textbf{Graph Neural Networks as Dynamical Systems.}
Adopting the interpretation of convolutional neural networks (CNNs) as discretizations of ODEs and PDEs  \cite{RuthottoHaber2018, chen2018neural, zhang2019linked} to GNNs, works like GODE \cite{zhuang2020ordinary}, GRAND \cite{chamberlain2021grand}, PDE-GCN\textsubscript{D} \cite{eliasof2021pde}, GRAND++ \cite{thorpe2022grand} and others, propose to view GNN layers as time steps in the integration of the non-linear heat equation, allowing to control the diffusion (smoothing) in the network, to understand oversmoothing \cite{nt2019revisiting,oono2020graph,cai2020note} in GNNs. Thus, works like \cite{chien2021adaptive, luan2020complete, luan2022revisiting, di2022graphGRAFF} propose to utilize a \emph{learnable} diffusion term, thereby alleviating oversmoothing. Other architectures like PDE-GCN\textsubscript{M} \cite{eliasof2021pde} and GraphCON \cite{rusch2022graph} propose to mix diffusion and oscillatory processes (e.g., based on the wave equation) to avoid oversmoothing by introducing a feature energy preservation mechanism. Nonetheless, as noted in \cite{rusch2023survey}, besides alleviating oversmoothing, it is also important to design GNN architectures with improved expressiveness. Recent examples of such networks are \cite{gravina2022anti} that propose an anti-symmetric GNN to alleviate over-squashing \cite{alon2021oversquashing}, and \cite{wang2022acmp, choi2022gread} that formulate a reaction-diffusion GNN to enable non-trivial pattern growth.
In this paper, we build on the properties of ADR PDEs, that in addition to modeling diffusive and reactive processes, also allow to capture advective processes such as the transportation of node features.

On another note, CNNs and GNNs are also used  to accelerate PDE solvers \cite{raissi2018deep, long2018pdenet, li2020multipole, brandstetter2022message, saadat2022neural}, as well as to generate \cite{sanchez2019hamiltonian} and compute \cite{belbute_peres_cfdgcn_2020}  physical simulations. In this paper, we focus on the view of GNNs as the discretization of ADR PDEs, rather than using GNNs to solve PDEs. 

\textbf{Advection on Graphs.} Advection is a term used in Physics to describe the transport of a substance in a medium. In the context of graphs, advection is used to express the transport of information (features) on the graph nodes. The underlying process of advection is described by a continuous PDE, and several graph discretization techniques  \cite{chapman2015advection, hovsek2019discrete} are available.
The advection operator has shown its effectiveness in classical (i.e., non-neural) graph methods, from blood-vessel simulations \cite{deepa2022graph}, to traffic flow prediction \cite{borovitskiy2021matern}. In this paper, we develop a neural advection operator that is combined with neural diffusion and reaction operators, called ADR-GNN.

\section{Method}
\label{sec:method}

In this section, we first describe the general outline of a continuous ADR system in Section \ref{sec:contADR}, and present its graph discrete analog, named \emph{ADR-GNN} in Section \ref{sec:graphDisc_ADR}. 
We discuss ADR-GNN components in detail in Sections \ref{sec:timeIntegration}-\ref{sec:graph_operators}.

\textbf{Notations.} We define a graph by $\mathcal{G} = (\mathcal{V}, \mathcal{E})$, where $\mathcal{V}$ is a set of $n$ nodes and $\mathcal{E} \subseteq \mathcal{V} \times \mathcal{V}$ is a set of $m$ edges. We denote the 1-hop neighborhood of the $i$-th node by $\mathcal{N}_i$, and the node features by $\bfU \in \mathbb{R}^{n \times c}$, where $c$ is the number of features. The symmetric graph Laplacian reads $\bfL = \bfD - \bfA$, and the symmetric normalized  Laplacian is given by $\hat{\bfL} = \bfD^{-\frac{1}{2}} \bfL \bfD^{-\frac{1}{2}}$, where $\bfD$ is the degree matrix.

\subsection{Continuous Advection-Diffusion-Reaction Systems}
\label{sec:contADR}
The continuous PDE that describes an ADR system is given by:
\begin{eqnarray}
    \label{eq:contADR}
    \frac{\partial U}{\partial t} =   \underbrace{\nabla \cdot (V U)}_{\rm Advection} + \underbrace{K \Delta U}_{\rm Diffusion} + \underbrace{f(U, X,\theta_r)}_{\rm Reaction},
\end{eqnarray}
where $ \ X \in \Omega, \ t \in [0,T]$, accompanied by initial conditions  ${U}(X, t=0)$ and boundary conditions. Here, $U(X, t) = [u_1(X, t), \ldots, u_c(X, t)]: \mathbb{R}^{\Omega \times [0,T]} \rightarrow \mathbb{R}^{c}$ is a density function, written as a vector of scalar functions $u_s(X, t) , \ s = 1,\ldots,c$, that depend on the initial  location $X$ and time $t$.  The spatial domain $\Omega$ can be  $\mathbb{R}^{d}$ or a manifold $\mathcal{M} \subseteq \mathbb{R}^{d}$. From a neural network perspective, $u_s$ is referred to as a channel, interacting with other channels. 
The left-hand side of Equation \eqref{eq:contADR} is a time derivative that represents the change in features in time, as discussed in Section \ref{sec:related}. The right-hand side includes three terms:
\begin{itemize}
    \item \textbf{Advection.} Here, $V$ denotes a velocity function that transports the density $U$ in space, and $\div$ is the divergence operator.
    \item \textbf{Diffusion.} We denote the continuous Laplacian operator by $\Delta$. The Laplacian is scaled with a diagonal matrix $K = {\rm diag}(\kappa_1, \ldots, \kappa_c) \in \mathbb{R}^{c\times c},  \ \ \kappa_i\ge 0$  of  non-negative diffusion coefficients, each independently applied to its corresponding channel in $U$.
    \item \textbf{Reaction.} Here, $f(U, X, \theta_{r})$ is a non-linear pointwise function parameterized by $\theta_r$.
\end{itemize}

\subsection{Advection-Diffusion-Reaction on Graphs}
\label{sec:graphDisc_ADR}
Equation \eqref{eq:contADR} is defined in the continuum.
We now use a graph ${\mathcal G}=(\mathcal{V}, \mathcal{E})$ to discretize $\Omega$. The nodes $\mathcal{{V}}$ can be regarded as a discretization of $X$, that is, the $i$-th node is located in $\bfX_i$, and the edges $\mathcal{E}$ represent the topology of $\Omega$. 
Then, the \emph{spatial}, graph discretization of Equation \eqref{eq:contADR} is:
\begin{subequations}
\label{eq:discDR}
\begin{align}
\label{eq:discDRa}
{\frac {d\bfU(t)}{d t}} &=    {\bf DIV} \left(\bfV\left( \bfU(t), t;\bftheta_a(t)\right) \bfU(t) \right) \\ \nonumber &  -\hat\bfL \bfU(t) \bfK(t;\bftheta_d(t)) + f(\bfU(t), \bfX, t; \bftheta_{r}(t)),\\
\label{eq:discDRb}
\bfU(0) &= \bfU^{(0)} = g_{\rm in}(\bfX, \bftheta_0). 
\end{align}
\end{subequations}
Here, $\bfU(t)\in\mathbb{R}^{n \times c}$ is a matrix that describes the node features at time $t$. The advection term depends on the velocity function $\bfV$ parameterized by learnable weights $\bftheta_a(t)$. The precise discretization of the advection operator ${\bf DIV} (\bfV\cdot)$ is discussed in Section \ref{sec:graph_operators}. The diffusion is discretized using the symmetric normalized Laplacian\footnote{In PDE theory, the Laplacian is a negative operator, while in graph theory it is positive. Therefore it is required to multiply $\hat\bfL$ by a negative sign  in Equation \eqref{eq:discDRa} compared to Equation \eqref{eq:contADR}. } $\hat{\bfL}$ that is scaled with 
 a diagonal matrix with non-negative learnable diffusion coefficients on its diagonal $\bfK(t;\bftheta_d(t)) = {\rm{diag}}({\rm{hardtanh}}(\bftheta_d(t), 0,1))\ge 0$, where ${\rm{hardtanh}}(\bftheta_d(t), 0,1)$ clamps each element in $\bftheta_{d}\in \mathbb{R}^{c}$ to be between 0 and 1.
 The reaction term  $f$ from Equation \eqref{eq:discDRa} is a pointwise non-linear function realized by a multilayer-perceptron (MLP) parameterized by learnable weights $\bftheta_r(t)$. To obtain initial node embedding $\bfU^{(0)} \in \mathbb{R}^{n \times c}$ from the input features $\bfX \in \mathbb{R}^{n \times c_{in}}$, we use a fully-connected layer $g_{in}$ in Equation \eqref{eq:discDRb}.

In this work we focus on static and temporal node-level tasks, and we note that typically, $c$, the number of hidden channels of $\bfU(T) \in \mathbb{R}^{n \times c}$, is different than $c_{out}$, the number of channels of the target output $\bfY \in \mathbb{R}^{n \times c_{out}}$. Therefore the output of neural network, $\tilde{\bfY}$, is given by 
\begin{eqnarray}
\label{yFromu}
\tilde{\bfY} = g_{\rm out}(\bfU(T), \bftheta_{\rm out}) \in    \mathbb{R}^{n \times c_{out}},
\end{eqnarray}
where  $g_{\rm out}$ is a fully-connected layer with weights $\bftheta_{\rm out}$.

\textbf{The qualitative behavior of ADR-GNN.}
The ADR-GNN model combines the learning of three powerful terms. Namely, the learned parameters are $\bftheta_a$ the advection parameters,  $\bftheta_{d}$ the diffusion parameters, and $\bftheta_r$  the reaction parameters. Therefore, an ADR-GNN layer can express and model various phenomena. For example, if we set $\bftheta_{d}(t) = 0$, then there is no diffusion in the system, and the method is dominated by advection and reaction. If on the other hand, one learns a very small advection (i.e., the learned $\bfV$, to be discussed later, tends to retain all features in place), then a reaction-diffusion oriented system is obtained. Similarly, other combinations of advection, diffusion, and reaction can be achieved, because of the learning of the parameters of the system. Thus, ADR-GNN can be adopted to solve a host of problems, depending on dynamics and patterns mandated by the data, as we show later in our experiments in Section \ref{sec:experiments}.

\subsection{From an ODE to a Graph Neural Network - Time Discretization of ADR-GNN}
\label{sec:timeIntegration}
Equation~\eqref{eq:discDR} \emph{spatially} discretizes the PDE in Equation \eqref{eq:contADR}, yielding an ODE defined on the graph. The \emph{time} discretization of
the ODE yields a sequential process that can be thought of as
layers of neural networks \cite{HaberRuthotto2017, chen2018neural, E2017}. That is, upon discrete time integration of Equation \eqref{eq:discDRa}, we replace the notion of time $t$ with  $l$ layers, and a step size $h$, that is a positive scalar hyperparameter.

While it is possible to use many ODE discretization methods (see, e.g., \cite{HaberRuthotto2017, ANODEV2, chen2018neural, chamberlain2021grand} and references within), in various
applications where an ADR system arises, from flow in porous media \cite{coats2000note}, to PDE-based image segmentation \cite{vese2002multiphase}, and multiphase flow \cite{JFNK2011}, an operator-splitting (OS) \cite{ascher2008numerical} is utilized. 
We therefore also use an OS time discretization for Equation \eqref{eq:discDRa}, that yields a graph neural ADR layer, summarized in Algorithm \ref{alg1}. Composing several neural ADR layers leads to ADR-GNN. We further discuss the properties of the OS approach in Appendix \ref{app:operatorSplitting}. The exact discretizations of the ADR terms are derived in Section \ref{sec:graph_operators}. 
\begin{algorithm}

\caption{Graph Neural Advection-Diffusion-Reaction Layer}\label{alg1}
\textbf{Input:} Node features $\bfU^{(l)} \in \mathbb{R}^{n\times c}$\\
\textbf{Output:} Updated node features $\bfU^{(l+1)} \in \mathbb{R}^{n \times c}$ \\
1: Advection: \\$\bfU^{(l+1/3)} = \bfU^{(l)} + h{\bf DIV} (\bfV( \bfU^{(l)}, t;\bftheta_a^{(l)}) \bfU^{(l)}).$ \\
2: Diffusion: \\ $\bfU^{(l+2/3)}={\rm{mat}}\left((\bfI + h \bfK(t;\bftheta_d^{(l)}) \otimes \hat{\bfL})^{-1} {\rm{vec}}(\bfU^{(l+1/3)}) \right).$ \\
3: Reaction: \\ $\bfU^{(l+1)} = \bfU^{(l+2/3)} + h f(\bfU^{(l+2/3)}, \bfU^{(0)}, t; \bftheta_{r}^{(l)}).$
\end{algorithm}

\subsection{Discretized Graph Operators}
\label{sec:graph_operators}
We now elaborate on the discretized graph operators utilized in our ADR-GNN, summarized in Algorithm \ref{alg1}. Besides the combination of the learnable advection, diffusion, and reaction terms, which, to the best of our knowledge, was not studied in the context of GNNs, the main innovation here is the \emph{consistent, mass preserving, and stable} discretization of the advection operator. 

\textbf{Advection.}
To define the graph discretized advection operator, we extend the non-learnable advection operator from \cite{chapman2015advection}, into a learnable, neural advection operator. Our advection operator transports node features based on learned directed edge weights (velocities) $\{(\bfV_{i\rightarrow j},\bfV_{j\rightarrow i})\}_{(i,j)\in\mathcal{E}}$, where each $\bfV_{i\rightarrow j},\bfV_{j\rightarrow i}\in\mathbb{R}^c$, such that $0\le \bfV_{i \rightarrow j} \le 1$. The notation ${i \rightarrow j}$ implies that the weight transfers features from the $i$-th to $j$-th node. We further demand that the outbound edge weights associated with every node, per channel, sum to 1, i.e., $    \sum_{j\in\mathcal{N}_i}\bfV_{i \rightarrow j} = 1$. 
This constraint suggests that a node can at most transfer the total of its features to other nodes. First, we define the discretized divergence from Equation \eqref{eq:discDRa}, that operates on the learned edge weights $\bfV$:
\begin{align}
    \label{eq:discDivergence}
    {\bf DIV}_i(\bfV \bfU) &= \textstyle{\sum_{j\in\mathcal{N}_i} \bfV_{j \rightarrow i} \odot \bfU_j  - \bfU_i \odot \sum_{j\in\mathcal{N}_i} \bfV_{i \rightarrow j}} \\ \nonumber &= \textstyle{\sum_{j\in\mathcal{N}_i} \bfV_{j \rightarrow i} \odot \bfU_j  - \bfU_i },
\end{align}
where $\odot$ is the elementwise Hadamard product. 
Then, the graph advection operator in Algorithm \ref{alg1} is:
\begin{align}
    \label{eq:advectiond}
    \bfU_i^{(l+1/3)}   
    &= \bfU^{(l)}_i + \textstyle{h{\bf DIV}_i(\bfV^{(l)}\bfU^{(l)})} \\ \nonumber  &= 
    \textstyle{\bfU_i^{(l)} + h \left(\sum_{j\in\mathcal{N}_i} \bfV_{j \rightarrow i}^{(l)} \odot \bfU_j^{(l)} - \bfU_i^{(l)} \right)}.
\end{align}
Namely, the updated node features are obtained by adding the $\bfV_{j \rightarrow i}$ weighted inbound node features, while removing the $\bfV_{i \rightarrow j}$ weighted outbound node features, and $h$ is a positive step size. The scheme in Equation \eqref{eq:advectiond} is the forward Euler discretization. We now show that the proposed graph neural advection operator is \emph{mass conserving},  \emph{stable} and \emph{consistent} \footnote{See stability definition and proofs in Appendix \ref{app:advectionProperties}.}, meaning that our advection operator is adherent to the continuous advection PDE \cite{leveque}.

\begin{lemma}
\label{lemma:massConservation} Define the mass of the graph node features $\bfU^{(l)} \in \mathbb{R}^{n \times c}$ as the scalar $\rho^{(l)} = \sum \bfU^{(l)}$. Then the advection operator in Equation \eqref{eq:advectiond} is mass conserving, i.e., $\rho^{(l+1/3)} = \rho^{(l)}$.
\end{lemma}

\begin{lemma}
    \label{lemma:Stability}
    The advection operator in Equation \eqref{eq:advectiond} is stable.
    \end{lemma}

To learn a \emph{consistent} advection operator, i.e., an operator that mimics the directional behavior of the advection in Equation \eqref{eq:contADR}, we craft an edge weight $\bfV$ mechanism, shown in Algorithm \ref{alg2}, that yields  direction-oriented weights, i.e., we ensure that 
$\bfV_{i\rightarrow j}\neq\bfV_{j\rightarrow i}$, unless they are zeroes. 
\begin{algorithm}
\caption{Learning directional edge weights.} \label{alg2}
\textbf{Input:} Node features $\bfU^{(l)} \in \mathbb{R}^{n\times c}$\\
\textbf{Output:} Edge weights $ \bfV_{i \rightarrow j}^{(l)},  \bfV_{j \rightarrow i}^{(l)} \in\mathbb{R}^c$ 
\begin{algorithmic}[1]
\State Compute edge features: \Statex $\bfZ_{ij}^{(l)} = {\rm{ReLU}}(\bfU_i^{(l)}{\bfA_{1}^{(l)}} +  \bfU_j^{(l)}\bfA_2^{(l)})\bfA_3^{(l)}.$ 
\Statex
$
    \bfZ_{ji}^{(l)} = {\rm{ReLU}}( \bfU_j^{(l)}\bfA_1^{(l)} +  \bfU_i^{(l)}\bfA_2^{(l)})\bfA_3^{(l)}.$
\State Compute relative edge features:  \Statex$\bfV_{i \rightarrow j}^{(l)} = {\rm ReLU}( \bfZ_{ij}^{(l)} - \bfZ_{ji}^{(l)})\bfA_4^{(l)}.$ \Statex  $  \bfV_{j \rightarrow i}^{(l)} = {\rm ReLU}( -\bfZ_{ij}^{(l)} + \bfZ_{ji}^{(l)})\bfA_4^{(l)}.$
 \State Normalize to obtain edge weights: \Statex $\bfV_{i \rightarrow j}^{(l)} \leftarrow \frac{\exp( \bfV_{i \rightarrow j}^{(l)})}{\sum_{k \in \mathcal{N}_i} \exp(\bfV_{i \rightarrow k}^{(l)})}.$  \Statex $ \bfV_{j \rightarrow i}^{(l)} \leftarrow \frac{\exp( \bfV_{j \rightarrow i}^{(l)})}{\sum_{k \in \mathcal{N}_j} \exp(\bfV_{j \rightarrow k}^{(l)})}.$

\end{algorithmic}
\end{algorithm}

Here, $\bftheta_{a}^{(l)} = \{ \bfA_1^{(l)}, \bfA_2^{(l)}, \bfA_3^{(l)}, \bfA_4^{(l)} \}$ are learnable fully connected layers, and the $\exp$ is computed channel-wise.
We note that the sign of $\bfZ_{ij}-\bfZ_{ji}$ is opposite than that of $-\bfZ_{ij}+\bfZ_{ji}$ in Algorithm \ref{alg2}. Hence, after the $\rm{ReLU}(\cdot)$ activation, one of the edge weights, either $\bfV_{i \rightarrow j}$ or $\bfV_{j \rightarrow i}$ is guaranteed to be equal to zero, and the other will be non-negative. This allows the architecture to create significant asymmetry in the edge weights $\bfV$, as also seen in Figure \ref{fig:transportExample}.

\textbf{Diffusion.}
To discretize the diffusion term from Equation \eqref{eq:discDRa}, both explicit and implicit time discretizations can be used \cite{ascher2008numerical}. An explicit forward Euler discretization yields the following layer:
\begin{equation}
    \label{eq:forwardEulerDiscretization}
    \bfU^{(l+2/3)} = \bfU^{(l+1/3)} - h \left(\hat\bfL \bfU^{(l+1/3)} \bfK^{(l)}\right). 
\end{equation}
However, an explicit scheme requires using a small step size $h>0$, as it is marginally stable \cite{ascher2008numerical}. 
We therefore harness an implicit scheme,
which guarantees the stability of the diffusion \footnote{See \cite{koto2008imex, haber2019imexnet, chamberlain2021grand} for details on implicit vs. explicit schemes for diffusion processes and in neural networks.}, and reads:

\begin{eqnarray}
    \label{implicit}
    \bfU^{(l+2/3)} = {\rm mat}\left((\bfI + h \bfK^{(l)}\otimes 
    \hat{\bfL})^{-1} {\rm vec}(\bfU^{(l+1/3)}) \right).
\end{eqnarray}
Here, $\otimes$ is the Kronecker product, ${\rm vec()}$ is a flattening operator, and ${\rm mat()}$ reshapes a vector to a matrix.
The computation of $\bfU^{(l+2/3)}$ requires the solution of a linear system, solved by conjugate gradients\footnote{We note that the matrix $\bfI + h \bfK_l\otimes \hat{\bfL}$ is  positive definite and invertible, because the identity matrix is positive definite, $h$ is positive, $\bfK_l$ is non-negative, and the graph Laplacian $\hat{\bfL}$ is positive semi-definite.} \cite{govl, ascher2008numerical}. In our experiments we found 5 iterations to be sufficient.

\textbf{Reaction.}
Our reaction term is realized using MLPs. Recent works showed that utilizing both additive and multiplicative MLPs yields improved performance \cite{Jayakumar2020Multiplicative, choi2022gread,ben2023exploring}. Hence, we define 
\begin{align}
   & \label{refun}
     f(\bfU^{(l+2/3)}, \bfU^{(0)}; \bftheta_{r}^{(l)}) 
=   \nonumber \sigma ( \bfU^{(l+2/3)} \bfR_{1}^{(l)} \\&  + {\rm{tanh}}(\bfU^{(l+2/3)}\bfR_{2}^{(l)}) \odot \bfU^{(l+2/3)} + \bfU^{(0)}\bfR_{3}^{(l)} ),
\end{align}
as our reaction term in Equation \eqref{eq:discDRa}. Here, $\bftheta_r^{(l)} = \{ \bfR_{1}^{(l)}, \bfR_{2}^{(l)}, \bfR_{3}^{(l)} \}$ are trainable fully-connected layers, and $\sigma$ is non-linear activation function (ReLU in our experiments), that can also be coupled with batch-normalization. This term is integrated via forward Euler as in Algorithm \ref{alg1}.

\section{Experimental Results}
\label{sec:experiments}

We demonstrate our ADR-GNN on two types of tasks on real-world datasets:  node classification, and spatio-temporal node forecasting. Architectures and training details are provided in Appendix \ref{appendix:architectures}, and the runtimes and complexity of ADR-GNN are discussed in Appendix \ref{app:complexity}. We use a grid search to select hyperparameters, discussed in Appendix \ref{appendix:hyperparams}.
Datasets details and statistics are reported in Appendix \ref{app:datasets}. 
Overall, we propose the two following ADR-GNN architectures:
\begin{itemize}
    \item ADR-GNN$_{S}$. Here we follow a similar approach to typical neural networks, where different weights are learned for each layer. From a dynamical system perspective, this can be interpreted as an unrolled ADR  iteration \cite{mardani2018neural}. This architecture is suitable for 'static' datasets that do not involve temporal information, such as Cora, and is specified in Appendix \ref{app:staticArch}.
    \item ADR-GNN$_T$. A time-dependent ADR-GNN for temporal datasets. Compared to ADR-GNN$_{S}$, it also utilizes temporal embedding, discussed in Appendix \ref{app:temporalADR}.
\end{itemize}

\subsection{Node Classification}
\textbf{Homophilic graphs.}
We experiment with Cora \cite{mccallum2000automating}, Citeseer \cite{sen2008collective}, and Pubmed \cite{namata2012query}  datasets. We use the 10 splits from \cite{Pei2020Geom-GCN:} with train/validation/test split ratios of $48 \%/32\%/20\%$, and report their average accuracy in Table \ref{table:homophilic_fully}. In Appendix \ref{app:stds} we also provide the accuracy standard deviation. 
As a comparison, we consider multiple recent methods, such as GCN \cite{kipf2016semi}, GAT \cite{velickovic2018graph}, Geom-GCN \cite{Pei2020Geom-GCN:}, APPNP \cite{klicpera2018combining}, JKNet \cite{jknet}, MixHop \cite{abu2019mixhop},  WRGAT\cite{Suresh2021BreakingTL}, GCNII \cite{chen20simple}, PDE-GCN \cite{eliasof2021pde}, NSD \cite{bodnar2022neuralsheaf}, H2GCN \cite{zhu2020beyondhomophily_h2gcn}, GGCN \cite{yan2021two}, C\&S \cite{huang2020combining}, DMP \cite{yang2021diverse}, GREAD \cite{choi2022gread}, LINKX \cite{lim2021large},  ACMII \cite{luan2022revisiting}, Ord. GNN \cite{song2023ordered}, and FLODE \cite{maskey2023a}. We see that our ADR-GNN\textsubscript{S} outperforms all methods on the Cora and Pubmed datasets, and achieves close (0.12\%  accuracy difference) to the best performing PDE-GCN on Citeseer.

\label{sec:nodeExperiments}
\begin{table*}[t]
\centering
\begin{minipage}[]{.35\textwidth}
  \setlength{\tabcolsep}{3pt}
  \center{
  \begin{tabular}{cccc}
    \toprule
    Method & Cora & Citeseer & Pubmed \\
    Homophily & 0.81 & 0.80 & 0.74 \\
    \midrule
    GCN  & 85.77 & 73.68 & 88.13  \\
    GAT & 86.37 & 74.32 & 87.62 \\
    GCNII\textsuperscript{$\dagger$} & 88.49  & 77.13 & \second{90.30}  \\
    Geom-GCN\textsuperscript{$\dagger$} & 85.27 & \third{77.99} & 90.05\\
    APPNP &  87.87 & 76.53 & 89.40 \\
    JKNet & 85.25  & 75.85  & 88.94 \\
    MixHop  & 87.61 & 76.26 & 85.31 \\
    WRGAT & 88.20 & 76.81 & 88.52 \\
    PDE-GCN\textsuperscript{$\dagger$} & \third{88.60}  & \first{78.48} & 89.93 \\
    GRAND & 87.36  & 76.46 & 89.02 \\
    GRAND++ & 88.15  & 76.57 & 88.50 \\
    NSD\textsuperscript{$\dagger$}  & 87.14  & 77.14   & 89.49 \\
    GGCN  & 87.95  & 77.14   & 89.15 \\
    H2GCN  & 87.87  & 77.11   & 89.49 \\
    C\&S & \second{89.05} & 76.22 & 89.74 \\
    GRAFF\textsuperscript{$\dagger$} & 88.01 & 77.30 & 90.04 \\
    DMP\textsuperscript{$\dagger$}  & 86.52 & 76.87 & 89.27 \\
    GREAD\textsuperscript{$\dagger$} & 88.57 & 77.60 & \third{90.23} \\
    LINKX  & 84.64 & 73.19 & 87.86 \\ 
    ACMII\textsuperscript{$\dagger$} & 88.25 & 77.12 & 89.71 \\
    Ord. GNN  & 77.31 &  90.15 & 88.37 \\
    FLODE & 78.07 &89.02 & 86.44 \\
    \midrule
    ADR-GNN\textsubscript{S} &  \first{89.43} &  \second{78.36} & \first{90.55}  \\
    \bottomrule
  \end{tabular}}
    \caption{Node accuracy ($ \%$) on homophilic datasets.
  }   \label{table:homophilic_fully}   

  \end{minipage}
  \hspace{1em}
\begin{minipage}[]{.5\textwidth}
\centering

  \begin{center}
   \setlength{\tabcolsep}{3pt}
  \begin{tabular}{ccccccc}
    \toprule
    Method & Squirrel & Film &  Chameleon & Cornell & Texas & Wisconsin \\
    Homophily & 0.22 & 0.22 & 0.23 & 0.30  & 0.11 & 0.21 \\
        \midrule
    GCN  & 23.96 & 26.86 &  28.18 &  52.70 & 52.16 & 48.92 \\
    GAT & 30.03 & 28.45 & 42.93 & 54.32 & 58.38 & 49.41
    \\
    GCNII\textsuperscript{$\dagger$} & 38.47 & 32.87 &   60.61  & 74.86 & 69.46 & 74.12 \\
    Geom-GCN\textsuperscript{$\dagger$}& 38.32 & 31.63 &  60.90 & 60.81 & 67.57 & 64.12 \\
    PDE-GCN\textsuperscript{$\dagger$} & -- & -- &   66.01 & \second{89.73} & \second{93.24}  &  \third{91.76} \\
    GRAND & 40.05 & 35.62 &  54.67 & 82.16 & 75.68 & 79.41 \\
    GRAND++ & 40.06 & 33.63 &  56.20 & 81.89 & 77.57 & 82.75 \\ 
    NSD\textsuperscript{$\dagger$}  &  56.34 & 37.79  & 68.68  &  86.49 & 85.95 & 89.41 \\
    GGCN  & 55.17 & \third{37.81} &  71.14 & 85.68  & 84.86  &  86.86 \\
    H2GCN  & 36.48 & 35.70 & 60.11 & 82.70  & 84.86  &  87.65 \\
    FAGCN & 42.59 & 34.87 & 55.22 & 79.19 & 82.43 & 82.94 \\
    GPRGNN & 31.61 & 34.63 & 46.58 & 80.27 & 78.38 & 82.94 \\
    GRAFF\textsuperscript{$\dagger$} & 59.01   & 37.11  & 71.38 & 84.05  & 88.38 & 88.83 \\
    DMP\textsuperscript{$\dagger$}  & 47.26 & 35.72 & 62.28 & \third{89.19} & 89.19 & \second{92.16} \\
    GREAD\textsuperscript{$\dagger$} & 59.22 & \second{37.90} & 71.38 & 87.03 & \third{89.73} & 89.41   \\
    ACMP-GCN & --  & -- & -- & 85.40 & 86.20 & 86.10  \\ 
    LINKX  & 61.81 &  36.10 & 68.42 & 77.84 & 74.60 & 75.49 \\
    G\textsuperscript{2}\textsuperscript{$\dagger$} & \third{64.26} &  37.30 & \third{71.40} & 87.30 &  87.57 & 87.84\\
    ACMII\textsuperscript{$\dagger$} & \second{67.40} & 37.09 & \second{74.76} & 86.49 & 88.38 & 88.43 \\
    Ord. GNN & 62.44 & 37.99 & 72.28 & -- & -- & -- \\
    FLODE & 64.23 & 37.16 &  73.60 & -- & -- & -- \\
    \midrule
    ADR-GNN\textsubscript{S} & \first{72.54} & \first{39.16}  & \first{79.91} & \first{91.89} & \first{93.61} & \first{93.46} \\
    \bottomrule
  \end{tabular}
\end{center}
  \caption{Node accuracy ($ \%$) on \emph{heterophilic} datasets. 
  }   \label{table:heterophilic_fully}

  \end{minipage}
\end{table*}

\textbf{Heterophilic graphs.} While our ADR-GNN offers competitive accuracy on homophilic datasets, as discussed in Section \ref{sec:related}, ADR systems are widely used to model non-smooth phenomena and patterns, as often appear in heterophilic datasets
by their definition \cite{Pei2020Geom-GCN:}. We therefore utilize 10 heterophilic datasets from various sources. In Table \ref{table:heterophilic_fully} we compare the average accuracy of our ADR-GNN\textsubscript{S} with recent GNNs on the Squirrel, Film, and Chameleon from \cite{musae}, as well as the Cornell, Texas and Wisconsin datasets from \cite{Pei2020Geom-GCN:}, using the 10-splits from \cite{Pei2020Geom-GCN:} we train/validation/test split ratios of $48 \% / 32\% / 20\%$. We include more comparisons and the accuracy standard deviation in Appendix \ref{app:stds}. In addition to the previously considered methods, we also compare with  FAGCN \cite{fagcn2021}, GraphCON \cite{rusch2022graph}, GPR-GNN \cite{chien2021adaptive}, GRAFF \cite{di2022graphGRAFF}, ACMP-GCN \cite{wang2022acmp}, and G\textsuperscript{2} \cite{rusch2022gradient}. We see that ADR-GNN\textsubscript{S} offers accuracy that is in line with recent state-of-the-art methods.
In addition, 
we evaluate ADR-GNN\textsubscript{S} on the Twitch-DE, deezer-europe, Penn94, and arXiv-year datasets from \cite{lim2021new, lim2021large} to further demonstrate the efficacy of our method, in Appendix \ref{app:additional}. 

\subsection{Spatio-Temporal Node Forecasting}
\label{sec:temporalExperiments}
Classical ADR models are widely utilized to predict and model spatio-temporal phenomena \cite{fiedler2003spatio, adam2004use}. We therefore now evaluate our temporal ADR-GNN\textsubscript{T}  on several spatio-temporal node forecasting datasets.
To this end, we harness the software package PyTorch-Geometric-Temporal 
 \cite{rozemberczki2021pytorch} that offers a graph machine learning pipeline for spatio-temporal graph tasks. In our experiments, we use the Chickenpox Hungary, PedalMe London, and Wikipedia Math datasets from \cite{rozemberczki2021pytorch}, as well as the traffic speed prediction datasets METR-LA \cite{jagadish2014big} and PEMS-BAY \cite{chen2001freeway}.

For the first three datasets, we follow the  incremental training
mode, mean-squared-error (MSE) loss, and testing procedure from \cite{rozemberczki2021pytorch}.  We report the performance of ADR-GNN\textsubscript{T} and other models, in terms of MSE, in Table \ref{tab:predictive_performance}. We compare with several recent methods, namely, DCRNN \cite{li2018diffusion}, GConv \cite{gconvlstm}, GC-LSTM \cite{gclstm}, DyGrAE \cite{dyggnn, dyngrae_1}, EGCN \cite{evolvegcn}, A3T-GCN \cite{a3tgcn}, T-GCN \cite{tgcn}, MPNN LSTM \cite{panagopoulos2020transfer},  AGCRN \cite{bai2020adaptive}, and DIFFormer \cite{wu2023difformer}. Our results in Table \ref{tab:predictive_performance} show improved performance, further revealing the significance of neural ADR systems on graphs.
\begin{table}[h]
\centering
{
\footnotesize
\resizebox{1\linewidth}{!}{
\begin{tabular}{ccccccc}
    \toprule
        \multirow{1}{*}{Method}
        & \multicolumn{1}{c}{{Chickenpox}}&  \multicolumn{1}{c}{{PedalMe}} & \multicolumn{1}{c}{Wikipedia} \\
\midrule
{DCRNN} &    1.124 $\pm$ 0.015            &       1.463$\pm$ 0.019     &        
\third{0.679 $\pm$ 0.020}        \\
{GConvGRU} & 1.128 $\pm$ 0.011             &      1.622 $\pm$ 0.032     &          \second{0.657 $\pm$ 0.015}          \\
{GC-LSTM} &    1.115 $\pm$ 0.014       &      \third{1.455 $\pm$ 0.023}     &     0.779 $\pm$  0.023           \\
{DyGrAE} & 1.120 $\pm$ 0.021            &      1.455 $\pm$ 0.031     &     0.773 $\pm$ 0.009          \\
{EGCN-O} &    1.124 $\pm$ 0.009       &    1.491 $\pm$ 0.024     &       
0.750 $\pm$ 0.014        \\
{A3T-GCN}& \third{1.114 $\pm$ 0.008}       &     1.469 $\pm$ 0.027  &      0.781 $\pm$ 0.011           \\
{T-GCN} & 1.117 $\pm$ 0.011    &    1.479 $\pm$ 0.012     &           0.764 $\pm$ 0.011      \\
{MPNN LSTM} & 1.116 $\pm$ 0.023           &       1.485 $\pm$ 0.028      &     0.795 $\pm$ 0.010         \\
{AGCRN} & 1.120 $\pm$ 0.010         &      1.469 $\pm$ 0.030   &    0.788 $\pm$ 0.011 \\
{DIFFormer} & 0.920 $\pm$ 0.001         &      --   &    0.720 $\pm$  0.036\\
\midrule
ADR-GNN\textsubscript{T} & \first{0.817 $\pm$ 0.012} & \first{0.598 $\pm$ 0.050} & \first{0.571 $\pm$ 0.014}
\\
\bottomrule
\end{tabular}
}
}
\caption{The performance of spatio-temporal networks evaluated by the average MSE of 10 experimental repetitions and standard deviations, calculated on 10\% forecasting horizons.}\label{tab:predictive_performance}
\end{table}

On the METR-LA and PEMS-BAY datsets, we follow the same training and testing procedures, and mean-absolute-error (MAE) loss as in \cite{li2018diffusion}. We report the MAE, root mean squared error (RMSE), and mean absolute percentage error (MAPE). To demonstrate the effectiveness of ADR-GNN\textsubscript{T} for varying time frame predictions, we report the results on 3, 6, and 12 future frame traffic speed prediction, where each time frame equates to 5 minutes. We compare ADR-GNN\textsubscript{T} with various methods, from 'classical' approaches such as historical averaging (HA), VAR \cite{lu2016integrating}, and SVR \cite{smola2004tutorial}, to neural methods like FC-LSTM \cite{sutskever2014sequence}, DCRNN \cite{li2018diffusion}, Graph WaveNet \cite{waveNet}, ASTGCN \cite{guo2019attention}, STSGCN \cite{song2020spatial}, GMAN \cite{zheng2020gman}, MTGNN  \cite{wu2020connecting}, GTS \cite{shang2021discrete}, STEP \cite{shao2022pre}, and STAEformer \cite{liu2023spatio}.  We find that our ADR-GNN\textsubscript{T} offers lower (better) metrics than the considered methods. For instance, on METR-LA, ADR-GNN\textsubscript{T} reduces the MAE achieved by the recent STEP method from 3.37 to 3.19. We provide the results on METR-LA in Table \ref{table:metrLA_pemsBay}, with additional comparisons as well as results on PEMS-BAY in Appendix \ref{app:spatioTemporal}.

\begin{table*}[t]
   \setlength{\tabcolsep}{3pt}
    \centering

    \label{tab:main}
    \begin{tabular}{ccccccccccc}

      \toprule
       \multirow{2}*{{Dataset}} &\multirow{2}*{{Method}} & \multicolumn{3}{c}{{Horizon 3}} & \multicolumn{3}{c}{{Horizon 6}}& \multicolumn{3}{c}{{Horizon 12}}\\ 
      \cmidrule(r){3-5} \cmidrule(r){6-8} \cmidrule(r){9-11}
      &  & MAE & RMSE & MAPE & MAE & RMSE & MAPE & MAE & RMSE & MAPE\\
      \midrule
      &FC-LSTM         & 3.44  & 6.30  & 9.60\%        & 3.77  & 7.23  & 10.09\%      & 4.37  & 8.69  & 14.00\% \\ 
      &DCRNN           & 2.77  & 5.38  & 7.30\%        & 3.15  & 6.45  & 8.80\%       & 3.60  & 7.60  & 10.50\% \\ 
      &STGCN           & 2.88  & 5.74  & 7.62\%        & 3.47  & 7.24  & 9.57\%       & 4.59  & 9.40  & 12.70\% \\ 
      {METR}&Graph WaveNet   & 2.69  & \third{5.15}  & 6.90\%        & 3.07  & 6.22  & 8.37\%       & 3.53  & 7.37  & 10.01\% \\
      {-LA}&ASTGCN          & 4.86  & 9.27  & 9.21\%        & 5.43  & 10.61 & 10.13\%      & 6.51  & 12.52 & 11.64\% \\  
      &STSGCN          & 3.31  & 7.62  & 8.06\%        & 4.13  & 9.77  & 10.29\%      & 5.06  & 11.66 & 12.91\% \\  
      &GMAN            & 2.80  & 5.55  & 7.41\%        & 3.12  & 6.49  & 8.73\%       & \third{3.44}  & 7.35  & 10.07\% \\  
      &MTGNN           & 2.69  & 5.18  & \third{6.88\%}        & 3.05  & \third{6.17}  & \third{8.19\%}       & 3.49  & \third{7.23}  & \third{9.87\%} \\  
      &GTS             & \third{2.67}  & 5.27  & 7.21\%        & \third{3.04}  & 6.25  & 8.41\%       & 3.46  & 7.31  & 9.98\% \\  
    &STEP      & \second{2.61}  & \second{4.98} & \second{6.60\%}        & \second{2.96}  & \second{5.97}  & \second{7.96\%}      & \second{3.37}  & \second{6.99}  & \second{9.61\%} \\ 
    & STAEformer   & 2.65 & 5.11 & 6.85\% & 2.97 & 6.00 & 8.13\% & 3.34 & 7.02 & 9.70\%
    \\
       \cmidrule(r){2-11}
    &   ADR-GNN\textsubscript{T} & \first{2.53} & \first{4.85} & \first{6.51\%} &  \first{2.81}  & \first{5.82} & \first{7.39\%}  & \first{3.19} & \first{6.89}  & \first{9.10\%} 
 \\
      \bottomrule
    \end{tabular}
        \caption{Multivariate time series forecasting on the METR-LA. Additional results are provided in Appendix \ref{app:spatioTemporal}.} \label{table:metrLA_pemsBay}
  \end{table*}
\subsection{Ablation Studies}
\label{sec:ablation}
\textbf{Synthetic Feature Transportation.} The benefit of diffusion and reaction are known in GNNs (see \cite{gasteiger_diffusion_2019, chamberlain2021grand, choi2022gread} and references within). However, the significance of neural advection was not studied in GNNs prior to our work, to the best of our knowledge. Therefore, and following the discussion of the task of feature transportation in Section \ref{sec:intro} and Figure \ref{fig:transportExample}, we now compare the behavior of the advection, diffusion, and reaction terms on this task. 
Although this experiment is conceptually simple, it is evident from Figure \ref{fig:syntheticExample}, that diffusion and reaction terms in GNNs are limited in modeling such a behavior.
This result, however, is not surprising. Employing diffusion smooths, rather than directly \emph{transferring} node features. Similarly, the reaction term can only learn to scale the node features in this experiment. On the contrary,  Figure \ref{fig:syntheticExample} that the advection term, that by definition, transports information, achieves an exact fit. More experimental details are given in Appendix \ref{app:synthetic}.

\begin{figure*}[h]
\centering
    \subfloat[Source]{    
\includegraphics[width=0.125\textwidth]{figs/source_cbar.eps}}
    \subfloat[Target]{  
\includegraphics[width=0.125\textwidth]{figs/target_cbar.eps}}
        \subfloat[Advection]{
\includegraphics[width=0.125\textwidth]{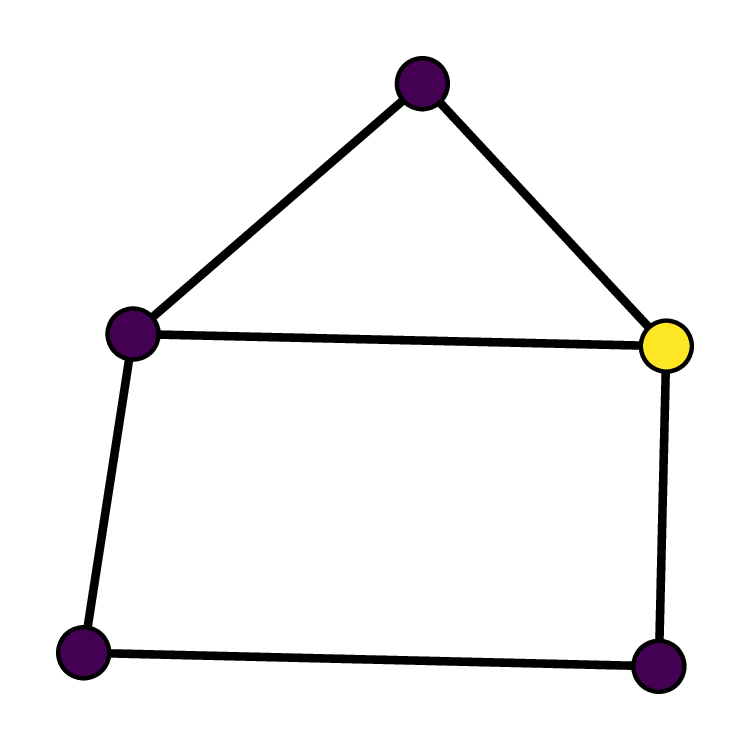}}
        \subfloat[Diffusion]{
\includegraphics[width=0.125\textwidth]{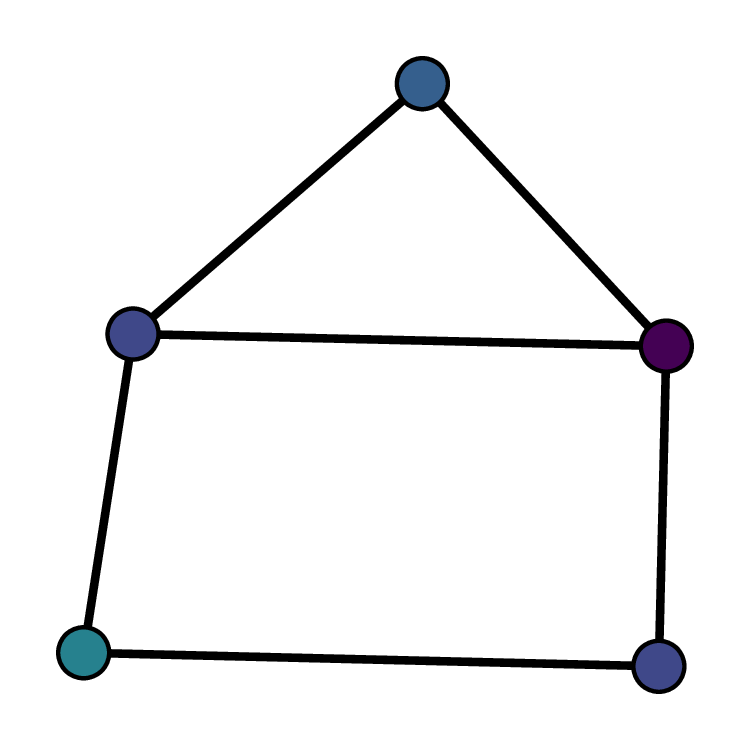}}
        \subfloat[Reaction]{
\includegraphics[width=0.125\textwidth]{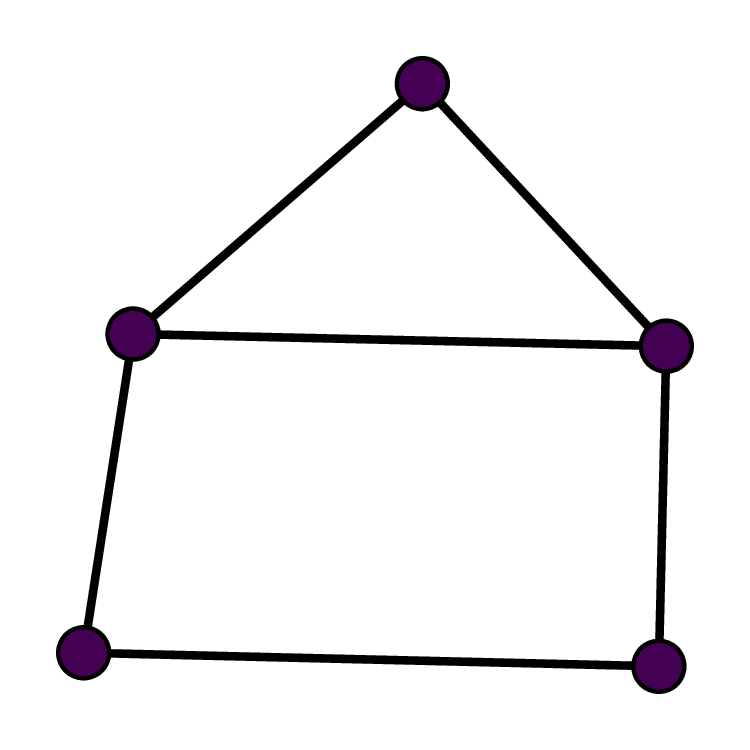}}
\includegraphics[width=0.045\textwidth,trim={0cm 0cm 7.5cm 0},clip]{figs/colorbar_adv.eps}
\caption{Source and target node features, and their fit using advection, diffusion, and reaction.}
\label{fig:syntheticExample}
\end{figure*}

\paragraph{The Impact of Advection, Diffusion, and Reaction.}
We study the influence of each of the proposed terms in Equation \eqref{eq:discDRa} on real-world datasets, independently and jointly. The results, reported in Table \ref{table:ADR_influence} further show the significance of the advection term. For datasets that are homophilic like Cora, we see minor accuracy improvement when incorporating the advection term. This is in line with known findings regarding the benefits of diffusion for homophilic datasets \cite{gasteiger_diffusion_2019, chamberlain2021grand}. More importantly, we see that mostly for heterophilic datasets like Chameleon, as well as traffic prediction datasets like PEMS-BAY, utilizing the advection significantly improves the performance of the network. Overall, we see that allowing all terms to be learned leads to favorable performance, indicating an implicit balancing of the terms obtained in the training stage.

\textbf{The Influence of Number of Layers.} 
The design of ADR-GNN can alleviate oversmoothing in two ways. First, by learning the diffusion coefficients $\bfK$, ADR-GNN controls the amount of smoothing, and can also achieve no smoothing if $\bfK$ is zero, depending on the data. Second, note that the advection and reaction terms can increase the frequency of the node features, because they are not limited to smoothing processes. 
To verify our observation, we evaluate ADR-GNN\textsubscript{S} on Cora and Citeseer with 2 to 64 layers, to see if its performance degrades as more layers are added, an issue that is associated with oversmoothing. We report the obtained accuracy in Figure
 \ref{fig:depth_study}, where no performance drop is evident. For reference, we also report the results obtained with GCN \cite{kipf2016semi}. Also, we define and report the measured Dirichlet energy in Appendix \ref{app:oversmoothing}, which shows that ADR-GNN does not oversmooth.

    \begin{table}[t]
    \centering
  \center{ 
  \begin{tabular}{ccccccc}
    \toprule
    \multirow{2}*{$\bfA$} & \multirow{2}*{$\bfD$} & \multirow{2}*{$\bfR$} & \multirow{2}*{Cora} &\multirow{2}*{Cham.} & METR & PEMS  \\
    & & & & & -LA & -BAY \\
    \midrule
   \cmark & \xmark &  \xmark & 86.69 & 66.79& 1.84 & 3.39 \\
       \xmark & \cmark &  \xmark & 88.21 & 65.08 & 1.93 & 3.67 \\
       \xmark & \xmark &  \cmark & 77.76 & 52.28 & 2.19 & 4.24 \\
         \cmark & \cmark &  \xmark & \third{88.92} & \third{73.33} & \third{1.79} & \third{3.30} \\
      \xmark & \cmark &  \cmark & \second{89.33} & 72.08 & 1.82 & 3.46 \\
      \cmark & \xmark &  \cmark & 88.02 & \second{73.46} & \second{1.71} & \second{3.21} \\
      \cmark & \cmark &  \cmark & \first{89.43} & \first{79.91} & \first{1.68} & \first{3.19} \\
    \bottomrule
  \end{tabular}}
      \captionof{table}{Impact of Advection ($\bfA$), Diffusion ($\bfD$), and Reaction ($\bfR$) on the Accuracy (\%) on Cora and Chameleon, MAE on METR-LA and PEMS-BAY.}
            \label{table:ADR_influence}  
    \end{table}

\begin{figure}[t]
    \centering
\begin{tikzpicture}
  \begin{axis}[
      width=1.0\linewidth, 
      height=0.5\linewidth,
      grid=major,
      grid style={dashed,gray!30},
      xlabel=Layers,
      ylabel=Accuracy (\%),
      ylabel near ticks,
      legend style={at={(0.35,0.32)},anchor=north,scale=0.8, draw=none, cells={anchor=west}, font=\small, fill=none},
      legend columns=2,
     xtick={0,4,8,16,32,64},
      xticklabels = {2,4,8,16,32,64},
          yticklabel style={
            /pgf/number format/fixed,
            /pgf/number format/precision=3
          },
      scaled y ticks=false,
      every axis plot post/.style={thick},
      ymin = 0
    ]
    \addplot[red, mark=oplus*, forget plot]
    table[x=nlayer,y=cora,col sep=comma] {data/depth_study.csv};
    \addplot[blue, mark=oplus*, forget plot]
    table[x=nlayer,y=citeseer,col sep=comma] {data/depth_study.csv};
    \addplot[red, style=dotted, mark=oplus*, forget plot]
    table[x=nlayer,y=cora_gcn,col sep=comma] {data/depth_study.csv};
    \addplot[blue, style=dotted, mark=oplus*, forget plot]
    table[x=nlayer,y=citeseer_gcn,col sep=comma] {data/depth_study.csv};
    \addplot[red, draw=none] coordinates {(1,1)};
    \addplot[blue, draw=none] coordinates {(1,1)};
    \addplot[gray, draw=none] coordinates {(1,1)};
    \addplot[gray, style=dotted, draw=none] coordinates {(1,1)};
    \legend{Cora, Citeseer, ADR-GNN\textsubscript{S}, GCN}
    \end{axis}
\end{tikzpicture}
\captionof{figure}{Accuracy (\%) vs. model depth. 
}
\label{fig:depth_study}
\end{figure}
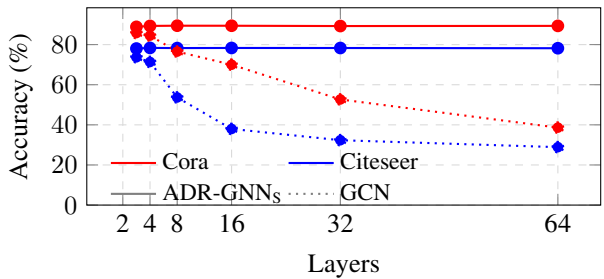

\section{Summary and Discussion}
\label{sec:summary}
In this paper, we present a novel GNN architecture that is based on the Advection-Diffusion-Reaction PDE, called ADR-GNN. We develop a graph neural advection operator that mimics the continuous advection operator, and compose it with learnable diffusion and reaction terms. 

We discuss and analyze the properties of ADR-GNN and its flexibility in modeling various phenomena. In particular, we show that the main advantage of the graph advection operator is its ability to transport information over the graph edges through the layers - a behavior that is hard to model using the diffusion and reaction terms that have been used in the literature. To demonstrate the effectiveness of ADR-GNN we experiment with total of 18 real-world datasets, from homophilic and heterophilic node classification to spatio-temporal node forecasting datasets.

While the gains observed on homophilic datasets are relatively modest, the performance improvements demonstrated on heterophilic datasets are significant, offering 5\% accuracy increase in some cases. Moreover, when applied to spatio-temporal node forecasting datasets, our ADR-GNN exhibits notable enhancements in the evaluated metrics compared to other methods. This progress can be attributed to the inherent suitability of ADR-GNN for tasks involving directional transportation of features, making it an intuitive choice for modeling such scenarios.

\section*{Acknowledgements}
This research was supported by grant no. 2018209 from the United States - Israel Binational
Science Foundation (BSF), Jerusalem, Israel, and by the Israeli Council for Higher Education (CHE) via the Data Science Research Center and the Lynn and William Frankel Center for Computer Science at
BGU. ME is supported by Kreitman High-tech
scholarship.

\bibliography{biblio}

\clearpage

\appendix
\section{The Behavior of the Operator Splitting Approach}
\label{app:operatorSplitting}
The advantage of Operator Splitting (OS) is that it allows the individual treatment of each component of the ODE separately, thus obtaining the appropriate qualitative behavior. This is especially beneficial in the context of advection, where it is difficult to obtain stability, that is, to satisfy the CFL condition \cite{leveque}, as well as to obtain mass conservation. Furthermore, OS is beneficial for the diffusion component, where implicit methods guarantee stability \cite{ascher2008numerical} compared to unstable explicit discretizations.

We now discuss the behavior of the OS approach for integrating the ADR ODE in Equation \eqref{eq:discDR}.
The theory behind OS can be analyzed in the linear case, and in the case of non-linear equations, linearization is typically assumed \cite{mattheij2005partial}. We now consider a linear ODE of the form
\begin{equation}
    \label{eq:osODE}
    \frac{d\bfU(t)}{dt} = \bfA \bfU + \bfD \bfU + \bfR \bfU,
\end{equation} 
where $\bfA \ , \bfD \ , \bfR$ denote the advection, diffusion, and reaction operators, respectively. $\bfU(t)$ denotes the node features at time $t$.

Suppose that we are interested in
computing $\bfU(t + \delta t)$. For this constant ODE system, the \emph{analytic}, exact solution is given by \cite{EvansPDE}:
\begin{equation}
    \label{eq:linearODE}
    \bfU(t + \delta t) = \exp\left( \delta t (\bfA + \bfD + \bfR) \right) \bfU(t).
\end{equation}
The following Lemma is easily proven using Taylor series of the matrix exponential function \cite{ascher2008numerical}:
\begin{lemma}
\label{lem:solutionError}
Let $\bfQ = \exp(\delta t (\bfA + \bfD + \bfR))$ where $\bfA$, $\bfD$, and $\bfR$ are matrices that do not share their eigenvectors.
Then the discrepancy between the exact and OS solution operator reads
$$\bfQ  - \exp(\delta t \bfR) \exp(\delta t \bfD)\exp(\delta t \bfA)  = {\cal O} (\delta t^2) $$
\end{lemma}

\begin{remark}
If the eigenvectors of $\bfA,\bfD,\bfR$ from Lemma \ref{lem:solutionError} are shared, then the matrix exponents commute, and the discrepancy is zero.
\end{remark}

Following Lemma \ref{lem:solutionError}, it holds that the solution of the ADR ODE can be expressed as a sequential process of three separate problems, with an error of ${\cal O}(\delta t^2)$ compared to the exact solution, as follows:
\begin{align}
&\exp\left( \delta t (\bfA + \bfD + \bfR) \right)\bfU(t) \\
&=\exp\left( \delta t \bfR \right)  \exp \left( \delta t\bfD \right) \exp\left( \delta t \bfA \right)\bfU(t) + {\cal O}(\delta t^2) \\
&=
 \underbrace{ \exp( \delta t \bfR )  \underbrace{ ( \exp\left( \delta t\bfD \right) \underbrace{\left(\exp\left( \delta t \bfA \right)\bfU(t) \right)}_{\rm Advection}}_{\rm Advection-Diffusion} }_{\rm Advection-Diffusion-Reaction}  + {\cal O}(\delta t^2)
\end{align}

Note that $\bfU^{(l+1/3)} = \exp(\delta t \bfA) \bfU^{(l)}$ is the exact solution \cite{hls} of the reaction system $\frac{\bfU(t)}{dt} = \bfA \bfU$. Similarly, $\bfU^{(l+2/3)} = \exp(\delta t \bfD) \bfU^{(l+1/3)}$
is the exact solution of the diffusion system $\frac{\bfU(t)}{dt} = \bfD \bfU$, and $\bfU^{(l+1)} = \exp(\delta t \bfR) \bfU^{(l+2/3)}$ is the exact solution of the reaction system $\frac{\bfU(t)}{dt} = \bfR \bfU$.
Thus, operator splitting can be viewed as taking a step of advection, followed by a step of diffusion that is finally followed by a reaction step. Note that the error above is of the same magnitude as each step of the forward Euler integration scheme, often used in neural networks, e.g., in ResNet (see [A] in additional references below). 

\section{The Properties of the Graph Discretized Advection Operator}
In this section, we prove Lemma \ref{lemma:massConservation} and Lemma \ref{lemma:Stability} from the main text. 
For convenience, we repeat the Lemmas, followed by their proofs. 
\label{app:advectionProperties} We start by noting the following remark:
\begin{remark}
    \label{remark:advectionChannelMixing} The advection operator in Equation \eqref{eq:advectiond} does not mix the node feature $\bfU$ channels in our ADR-GNN.
\end{remark}
The importance of this remark is that it allows us to analyze the properties of the advection operator per-channel, or, alternatively, assuming a single channel, $c=1$, which we assume in our following proofs.

\begin{lemma_app}
\label{lemma:massConservationAppendix} Define the mass of the graph node features $\bfU^{(l)} \in \mathbb{R}^{n \times c}$ as the scalar $\rho^{(l)} = \sum \bfU^{(l)}$. Then the advection operator in Equation \eqref{eq:advectiond} is mass conserving, i.e., $\rho^{(l+1/3)} = \rho^{(l)}$.
\end{lemma_app}
\begin{proof}
Without loss of generality, and following Remark \ref{remark:advectionChannelMixing}, let us assume a single channel, and consider the mass of the node features  $\rho=\sum\bfU$, before and after applying an advection layer as described in Equation \eqref{eq:advectiond}. The input node features has a total mass of $\rho^{(l)} = \sum\bfU^{(l)}$. The total mass of the output of an advection layer reads:
\begin{subequations}
\small
\label{eq:proofEqs}    
\begin{align}
    \label{eq:totalSumPriorAdvection}
    &\rho^{(l+1/3)} \\  =&\sum_{i}\bfU_{i}^{(l+1/3)}\\ =&  \sum_{i} \left(\bfU_{i}^{(l)} + h\sum_{j \in \mathcal{N}_i} \bfV_{j \rightarrow i}^{(l)} \bfU_j^{(l)} - h\bfU_i^{(l)} \right)
    \label{eq:transition_before_all_J}\\  
     =&  \sum_{i} \left(\bfU_{i}^{(l)} + h\sum_{j} \bfV_{j \rightarrow i}^{(l)} \bfU_j^{(l)} - h\bfU_i^{(l)} \right)      \label{eq:transition_withAll_J} \\  
     =& \sum_{i}\bfU_{i}^{(l)} +  h\left(\sum_{i}\sum_{j} \bfU_j^{(l)} \left( \bfV_{j \rightarrow i}^{(l)}\right)  - \sum_{i} \bfU_i^{(l)}\right)  \\ 
     \label{eq:transitionOrder}
     =& \sum_{i}\bfU_{i}^{(l)} +  h\left(\sum_{j} \bfU_j^{(l)} \left(\sum_{i} \bfV_{j \rightarrow i}^{(l)}\right)  - \sum_{i} \bfU_i^{(l)}\right) \\ 
     =& \sum_{i}\bfU_{i}^{(l)} +  h\left(\sum_{j} \bfU_j^{(l)}  - \sum_{i} \bfU_i^{(l)}\right) \\ =& \sum_{i}\bfU_{i}^{(l)} = \rho^{(l)}.
     \label{eq:transitionSum}
\end{align}
\end{subequations}

The transition between Equations \eqref{eq:transition_before_all_J} and \eqref{eq:transition_withAll_J}  is valid because for $j \notin \mathcal{N}_i$, the edge weight is zero, i.e., $\bfV_{j \rightarrow i}^{(l)} = 0$, and therefore the summation does not change.
Also, the transition between Equations \eqref{eq:transitionOrder} and \eqref{eq:transitionSum} holds because of the constraint on the sum of outbound edge weights to be equal to $1$, i.e. $\sum_{i \in \mathcal{N}_j} \bfV_{j \rightarrow i}^{(l)} = 1$.
Therefore, because $\rho^{(l+1/3)} = \rho^{(l)}$, our graph discretized advection operator is mass preserving. 
\end{proof}

\begin{definition}
    \label{def:stability}
    A neural operator $F$ that considers node features $\bfU$ is (Lyapunov) \emph{stable} if for every $\epsilon > 0$ there exists $\delta > 0$, such that every pair of inputs $\bfU \ , \tilde{\bfU}$ that satisfy $\| \bfU - \tilde{\bfU}\| \leq \delta$, then $\| F(\bfU) - F(\tilde{\bfU}) \| \leq \epsilon$.
\end{definition}

\begin{lemma_app}
    \label{lemma:StabilityAppendix}
    The advection operator in Equation \eqref{eq:advectiond} is stable.
\end{lemma_app}
\begin{proof}
    Without loss of generality, and following Remark \ref{remark:advectionChannelMixing}, let us consider a single channel, and let $\bfV^{(l)}$ be a sparse matrix such that $\bfV_{ij}^{(l)} = \bfV_{i\rightarrow j}^{(l)}$. To show stability, we first observe that the matrix form of the scalar formulation of the advection \emph{layer} in Equation \eqref{eq:advectiond} is given by:
    \begin{align}
        \label{eq:MatrixForm}
        \bfU^{(l+1/3)} &=& \bfI \bfU^{(l)} + h\bfV^{(l)}\bfU^{(l)} - h\bfU^{(l)} \\ &=& \underbrace{\left((1-h)\bfI + h\bfV^{(l)}\right)}_{\bfA^{(l)}}\bfU^{(l)}.
    \end{align}
    Because of the demand that $\bfV^{(l)}$ is normalized (i.e., $\sum_{j\in \mathcal{N}_i}\bfV_{i \rightarrow j}^{(l)} = 1$), and the advection weights satisfy $0 \leq \bfV_{i\rightarrow j}^{(l)} \leq 1$, the advection operator $\bfA^{(l)}$ is a column stochastic non-negative matrix. By the Perron-Frobenius theorem, such matrices are known to have a spectral radius bounded by 1 (see [B] in additional appendix references), and hence the advection operator is stable. 
\end{proof}

\section{Architectures and Training Details}
\label{appendix:architectures}
As discussed in the main paper, we propose two architectures, depending on the type of dataset - static (e.g., Cora), or spatio-temporal (e.g., PEMS-BAY). In the following subsections, we elaborate on these architectures.
\subsection{Node Classification: ADR-GNN\textsubscript{S}}
\label{app:staticArch}
We now elaborate on the `static' architecture ADR-GNN\textsubscript{S} used in our node classification experiments. The overall architecture is similar to standard GNN architectures for node classification, such as GCN \cite{kipf2016semi} and GCNII \cite{chen20simple}. It is composed of an initial embedding layer (that corresponds to Equation \eqref{eq:discDRb} in the main paper), $L$ graph neural ADR layers, and a classifier, as described in Equation \eqref{yFromu} in the main paper. The complete flow of ADR-GNN\textsubscript{S} is described in Algorithm \ref{alg:staticADR}. To train ADR-GNN\textsubscript{S} on node classification datasets, we minimize the cross-entropy loss between the ground-truth node labels $\bfY$ and the predicted node labels $\tilde{\bfY}$, as is standard in GNNs, and similar to \cite{kipf2016semi}.

    \begin{algorithm*}[t]
    \caption{ADR-GNN\textsubscript{S} Architecture Flow}    \label{alg:staticADR}
    \hspace*{\algorithmicindent} \textbf{Input:} Node features $\bfX \in \mathbb{R}^{n \times c_{in}}$\\
    \hspace*{\algorithmicindent} \textbf{Output:} Predicted node labels $\tilde{\bfY}\in \mathbb{R}^{n \times  c_{out}}$ 
    \begin{algorithmic}[1]
    \Procedure{ADR-GNN\textsubscript{S}}{}
         \State $\bfX \gets {\rm{Dropout}}(\bfX, p)$

        \State $ \bfU^{(0)} = g_{in}(\bfX)$
    \For {$l = 0 \ldots L-1$}
       \State $\bfU^{(l)} \gets {\rm{Dropout}}(\bfU^{(l)}, p)$
\State Advection: $\bfU^{(l+1/3)} = \bfU^{(l)} + h{\bf DIV} (\bfV( \bfU^{(l)};\bftheta_a^{(l)}) \bfU^{(l)})$ 
\State Diffusion: $\bfU^{(l+2/3)} = {\rm mat}\left((\bfI + h \bfK(\bftheta_d^{(l)}) \otimes \hat{\bfL})^{-1} {\rm vec}(\bfU^{(l+1/3)}) \right)$
\State Reaction: $\bfU^{(l+1)} = \bfU^{(l+2/3)} + h f(\bfU^{(l+2/3)}, \bfU^{(0)}; \bftheta_{r}^{(l)})$
    \EndFor
         \State $\bfU^{(L)} \gets {\rm{Dropout}}(\bfU^{(L)}, p)$

    \State $\tilde{\bfY} = g_{out}(\bfU^{(L)})$
    \State Return $\tilde{\bfY}$
    \EndProcedure
    \end{algorithmic}
    \end{algorithm*}

\subsection{Spatio-Temporal Node Forecasting: ADR-GNN\textsubscript{T}}
\label{app:temporalADR}
The typical task in spatio-temporal datasets is to predict future quantities (e.g., driving speed) given several previous time steps (also called frames). Formally, one is given an input tensor $\bfX_{temporal} \in \mathbb{R}^{n \times 
 \tau_{in}  c_{in}}$, where $\tau_{in}$ is the number of input (observed) time frames, and the goal is to predict $\tau_{out}$ time frames ahead, i.e., $\bfY_{\rm{temporal}} \in \mathbb{R}^{n \times \tau_{out}  c_{out}}$. This is in contrast to 'static' datasets such as Cora \cite{mccallum2000automating}, where input node features $\bfX \in \mathbb{R}^{n \times c_{in}}$ are given, and the goal is to fit to some ground-truth $\bfY \in \mathbb{R}^{n \times c_{out}}$. In this context, a 'static' dataset can be thought of as setting $\tau_{in} = \tau_{out} = 1$ for the spatio-temporal settings.  We show the overall flow of our `temporal' architecture ADR-GNN\textsubscript{T} in Algorithm \ref{alg:temporalADR} \footnote{In Algorithm \ref{alg:temporalADR}, $\oplus$ denotes channel-wise concatenation.}. 

In our spatio-temporal ADR-GNN\textsubscript{T}, we update the hidden state feature matrix $\bfU^{(l)}_{\rm{state}}$ based on the hidden historical feature matrix $\bfU^{(l)}_{\rm{hist}}$, as shown in Lines 6-9 in Algorithm \ref{alg:temporalADR}.

Similarly to Attention models (see [C] in the additional appendix references), we incorporate time embedding based on the concatenation of sine and cosine function evaluations with varying frequencies multiplied by the time of the input frames, as input to our ADR-GNN\textsubscript{T}, denoted by ${\bfT_{\rm{emb}}} \in \mathbb{R}^{n \times \tau_{in} c_t }$, where we choose the number of frequencies to be 10, and by the concatenation of both sine and cosine lead to $c_t = 20$. We note that the time embedding is computed in a pre-processing fashion.
To initialize the hidden feature matrices $\bfU^{(0)}_{\rm{state}}, \ \bfU^{(0)}_{\rm{hist}}$, we embed the input data $\bfX_{\rm{temporal}}$, concatenated with ${\bfT_{\rm{emb}}}$, using two fully connected layers, as described in Lines 3-4 in Algorithm \ref{alg:temporalADR}. \footnote{In Python notations, $\bfX_{\rm{temporal}}[:,-c_{in}]$ extracts the last $c_{in}$ entries of the second dimension of $\bfX_{\rm{temporal}}$, which returns the features of the last time frame.}

For the Chickenpox Hungary, PedalMe London, and Wikipedia Math datasets, we minimize the mean squared error (MSE) between the ground truth future node quantities and the predicted quantities by ADR-GNN\textsubscript{T}, similar to the training procedure of the rest of the considered methods in Table \ref{tab:predictive_performance}. Specifically, following \cite{rozemberczki2021pytorch}, the goal is to predict the node quantities of the next time frame given 4 previous time frames. On the METR-LA and PEMS-BAY datasets we minimize the mean absolute error (MAE), similar to \cite{li2018diffusion}, where we also follow the standard 12 previous time frames as inputs, and consider 3,6, and 12 future time frames node quantity prediction as output.

    \begin{algorithm*}[t]
    \caption{ADR-GNN\textsubscript{T} Architecture Flow}    \label{alg:temporalADR}
    \hspace*{\algorithmicindent} \textbf{Input:} Node features $\bfX_{\rm{temporal}} \in \mathbb{R}^{n \times \tau_{in} c_{in}}$, time embedding ${\bfT_{\rm{emb}}} \in \mathbb{R}^{n \times \tau_{in}c_t}$  \\
    \hspace*{\algorithmicindent} \textbf{Output:} Predicted future node quantities $\tilde{\bfY}\in \mathbb{R}^{n \times \tau_{out}  c_{out}}$ 
    \begin{algorithmic}[1]
    \Procedure{ADR-GNN\textsubscript{T}}{}
        \State $\bfX_{\rm{temporal}} \gets {\rm{Dropout}}(\bfX_{\rm{temporal}}, p)$
    \State $\rm{{\bfT_{\rm{emb}}}} \gets  \rm{g^{\rm{time-embed}}}({\bfT_{\rm{emb}}})$
    \State $ \bfU^{(0)}_{\rm{state}} = g_{in}^{\rm{state}}(\bfX_{\rm{temporal}}[:,-c_{in}] \oplus {\bfT_{\rm{emb}}})$
        \State $ \bfU^{(0)}_{\rm{hist}} = g_{in}^{\rm{hist}}(\bfX_{\rm{temporal}} \oplus {\bfT_{\rm{emb}}})$
    \For {$l = 0 \ldots L-1$}    
    \State $\bfU_{\rm{state}}^{(l)} \gets {\rm{Dropout}}(\bfU_{\rm{state}}^{(l)}, p)$
    
    \State Advection: $\bfU_{\rm{state}}^{(l+1/3)} = \bfU_{\rm{state}}^{(l)} + h{\bf DIV} (\bfV( \bfU^{(l)}_{\rm{hist}};\bftheta_a^{(l)}) \bfU^{(l)}), \bftheta_a) \bfU_{\rm{state}}^{(l)}$ 
    \State Diffusion: $\bfU_{\rm{state}}^{(l+2/3)} = {\rm mat}\left((\bfI + h \bfK^{(l)}\otimes \hat{\bfL})^{-1} {\rm vec}(\bfU_{\rm{state}}^{(l+1/3)}) \right)$
    \State Reaction: $\bfU_{\rm{state}}^{(l+1)} = \bfU_{\rm{state}}^{(l+2/3)} + h f(\bfU_{\rm{hist}}^{(l+2/3)}, \bfU_{\rm{hist}}^{(0)}; \bftheta_r)$   
    \State $\bfU_{\rm{hist}}^{(l+1)} = g_{l}^{\rm{hist}}(\bfU_{\rm{hist}}^{(l)} \oplus \bfU_{\rm{state}}^{(l+1)} \oplus {\bfT_{\rm{emb}}})$
    \EndFor
    \State $\bfU_{\rm{state}}^{(L)} \gets {\rm{Dropout}}(\bfU_{\rm{state}}^{(L)}, p)$
    
    \State $\tilde{\bfY} = g_{out}^{\rm{state}}(\bfU_{\rm{state}}^{(L)})$
    \State Return $\tilde{\bfY}$
    \EndProcedure
    \end{algorithmic}
    \end{algorithm*}

\section{Computational Complexity and Time}
\label{app:complexity}
\textbf{Complexity.}
Our ADR-GNN architectures include four main operations: (i) input/output embedding, (ii) advection, (iii) diffusion, and (iv) reaction layers.

The complexity of (i) and (iv) is $ {\cal O}({|\mathcal{V}|}c^2)$, where $c$ is the number of channels, because they are composed of pointwise MLPs.
The complexity of (ii) is ${\cal O}((|\mathcal{V}|+ |\mathcal{E}|)c^2)$ because it requires the computation of the edge weights $\bfV$, as shown in Algorithm \ref{alg2}, followed by a multiplication by the node features, as shown in Equation \eqref{eq:advectiond}. Similarly, (iii) also is of complexity ${\cal O}((|\mathcal{V}| + |\mathcal{E}|)c)$, because it is required to multiply the scaled Laplacian with the node features. Note that as discussed in Section \ref{sec:graph_operators}, and similar to \cite{chamberlain2021grand, rusch2022graph}, we do not explicitly invert the matrix $\bfI + h \bfK^{(l)} \otimes \hat{\bfL}$, but rather use the conjugate-gradients (CG) method to solve a system of equations. Below, we further discuss the use of CG to solve the system of equations.

\textbf{Implicit solution of the diffusion term.}
The diffusion step in our ADR-GNN, as shown in Section \ref{sec:graph_operators}, requires the solution of the linear system at each step. As previously discussed this is solved by using the CG method.
Thus, the backward (differentiation) function in most common software packages, such as PyTorch, tracks the CG iterations. This tracking can be avoidable by using implicit differentiation, which is the backbone of implicit methods (see \cite{haberBook2014} for detailed derivation). In the context of deep learning, implicit differentiation was used for
implicit neural networks \cite{gu2020implicit}.
The basic idea is to use implicit differentiation
of the equation 
\begin{equation}
\bfU^{(l+2/3)} = {\rm{mat}}(((\bfI - h \bfK^{(l)} \otimes \hat\bfL ) {\rm{vec}}(\bfU^{(l+1/3)}))     
\end{equation}

with respect to  ${\bfK^{(l)}}$  and thus avoid the tracking of the CG iterations if many are needed.

\textbf{Runtimes.} In addition to the complexity analysis above, we  provide the measured runtimes in Table \ref{tab:runtimes}. Learning the advection weights requires an increased computational effort. However, it can significantly improve the considered task metric. For convenience, in Table \ref{tab:runtimes}, in addition to the runtimes we also report the obtained task metric. Importantly, we show that the improved metrics offered by ADR-GNN\textsubscript{S} with 64 channels and 4 layers are not simply obtained due to the increased costs, by showing that enlarging GCN and GAT from standard 2 layers and 64 channels, to 2 layers and 256 channels (wide), or 64 layers and 64 channels (deep) does not yield similar improvements. We measure the runtimes using an Nvidia-RTX3090 with 24GB of memory, which is the same GPU used to conduct our experiments.

\begin{table*}[t]
  \caption{Training and inference GPU runtimes (milliseconds), number of parameters (thousands), and node classification accuracy (\%) on Cora.}
  \label{tab:runtimes}
  \begin{center}
  \begin{tabular}{lcccccccccc}
  \toprule
   \multirow{2}{*}{Metric}  & \multirow{2}{*}{GCN} & \multirow{2}{*}{GAT} & GCN  & GAT& GCN & GAT    & \multirow{2}{*}{ADR-GNN\textsubscript{S}} \\
   & & & (wide) & (wide) & (deep) & (deep) &  \\
    \midrule
    Training time & 7.71 & 14.59  & 14.32 & 36.63 & 95.11 & 184.51 & 35.41 \\
    Inference time  & 1.75 & 2.98   &  2.86 & 7.57 & 12.93 &  38.96 & 8.24 \\
    Parameters & 104 & 105 & 565 & 567 & 358 & 360 & 210 \\
    Accuracy &  85.77 & 83.13  & 85.18 & 83.37 & 38.62 & 33.40  & 89.43 \\
    \bottomrule
  \end{tabular}
  \end{center}
\end{table*}

\section{Hyperparameters}
\label{appendix:hyperparams}
All hyperparameters were determined by grid search, and the ranges and sampling mechanism distributions are provided in Table \ref{tab:hyperparams}.
Note, that as discussed after Equation \eqref{refun}, we may add a BatchNorm layer before applying the non-linear activation $\sigma$ to the reaction term, we therefore treat the use of batchnorm as a hyperparameter in Table \ref{tab:hyperparams}.
\begin{table*}[t]
\centering
\caption{Hyperparameter ranges}
{
\label{tab:hyperparams}
\begin{tabular}{ccc}
\toprule
Hyperparameter   & Range & Uniform Distribution \\
\midrule
     input/output embedding learning rate &  [1e-4, 1e-1] & log uniform  \\
     advection learning rate &  [1e-4, 1e-1] & log uniform \\
    diffusion learning rate & [1e-4, 1e-1] & log uniform   \\
      reaction learning rate &  [1e-4, 1e-1] & log uniform  \\
       input/output embedding weight decay &  [0, 1e-2] & uniform \\
       advection weight decay & [0, 1e-2] & uniform  \\
    diffusion weight decay & [0, 1e-2] & uniform  \\
      reaction weight decay & [0, 1e-2] & uniform  \\
      input/output dropout &  [0, 0.9] & uniform \\
      hidden layer dropout & [0, 0.9] & uniform \\
      use BatchNorm & \{ yes / no \} & discrete uniform \\
      step size h &  [1e-3, 1] & uniform \\
      layers & \{ 2,4,8,16,32,64 \} & discrete uniform \\
      channels &  \{ 8,16,32,64,128,256 \} & discrete uniform  \\
 \bottomrule
\end{tabular}
}
\end{table*}

\section{Datasets}
\label{app:datasets}
We report the statistics of the datasets used in our experiments in Table \ref{table:datasets} and \ref{tab:desc_discrete} for the node classification, and spatio-temporal node forecasting datasets, respectively.
All datasets are publicly available, and appropriate references to the data sources are provided in the main paper.

\begin{table*}[t]
  \caption{Node classification datasets statistics.}
  \label{table:datasets}
  \begin{center}
  \begin{tabular}{lccccc}
  \toprule
    Dataset & Classes & Nodes & Edges & Features & Homophily \\
    \midrule
    Cora & 7 & 2,708 & 5,429 & 1,433 & 0.81\\
    Citeseer & 6 & 3,327  & 4,732 & 3,703 & 0.80\\
    Pubmed & 3 & 19,717 & 44,338 & 500 & 0.74 \\
    Chameleon & 5 & 2,277 &  36,101 & 2,325 & 0.23\\
    Film & 5 & 7,600 & 33,544 & 932 & 0.22  \\
    Squirrel & 5 & 5,201 & 198,493 &  2,089 & 0.22 \\
    Cornell & 5 & 183 & 295 & 1,703 & 0.30\\
    Texas & 5 & 183 & 309 & 1,703 & 0.11 \\
    Wisconsin & 5 & 251 & 499 & 1,703 & 0.21 \\
    Twitch-DE & 2 & 9,498 & 76,569 & 2,545 & 0.63 \\
    Deezer-Europe & 2 &  28,281 &  92,752 & 31,241  & 0.52  \\
    Penn94 (FB100) &  2 &  41,554 &  1,362,229 & 5 & 0.47  \\
    arXiv-year & 5 & 169,343 &  1,166,243 & 128 & 0.22 \\
    OGBN-Arxiv & 40 & 169,343 &  1,166,243 & 128 & 0.65 \\

    \bottomrule
  \end{tabular}
\end{center}
\end{table*}

\begin{table*}[t]
\centering
\caption{Attributes of the spatio-temporal datasets used in  \ref{sec:temporalExperiments} and information about the number of time periods ($T$) and spatial units ($|\mathcal{V}|$).}\label{tab:desc_discrete}
{
\begin{tabular}{cccc}
\toprule
Dataset   & Frequency & $T$ & $|\mathcal{V}|$ \\
\midrule
    Chickenpox Hungary & Weekly & 522 & 20 \\
     Pedal Me Deliveries &  Weekly & 36 & 15 \\
 Wikipedia Math & Daily & 731 & 1,068 \\
 METR-LA &   5-Minutes &  34,272 & 207 \\
 PEMS-BAY & 5-Minutes & 52,116 & 325 \\

 \bottomrule
\end{tabular}
}
\end{table*}

\section{Experimental Results}
\label{appendix:results}

\subsection{Additional Comparisons and Standard Deviations on Node Classification}
\label{app:stds}
To allow a more comprehensive comparison, and because some of the considered methods did not report the standard deviation around the mean accuracy, we now provide the experimental results from \ref{sec:nodeExperiments} on Cora, Citeseer, Pubmed datasets in Table \ref{table:homophilic_fully_std}, and Cornell, Texas, Wisconsin, Squirrel, Film, Chameleon datasets in Table \ref{table:heterophilic_fully_std} ,with the standard deviation around the mean of the 10 splits from \cite{Pei2020Geom-GCN:}. Note, that here we do not color the tables, because some of the second or third top performing models in the main paper did not report the accuracy standard deviation, and therefore coloring Tables \ref{table:homophilic_fully_std}-\ref{table:heterophilic_fully_std} will change the order of the best performing models.

\begin{table*}[t]
  \caption{Node classification accuracy ($ \%$) on \emph{homophilic} datasets. $\dagger$ denotes the maximal accuracy of several proposed variants.}
  \label{table:homophilic_fully_std}   
  \center{
  \begin{tabular}{cccc}
    \toprule
    Method & Cora & Citeseer & Pubmed \\
    Homophily & 0.81 & 0.80 & 0.74 \\
    \midrule
    GCN  & 85.77 $\pm$ 1.27 & 73.68 $\pm$ 1.36 & 88.13 $\pm$ 0.50  \\
    GAT & 86.37 $\pm$ 0.48 & 74.32 $\pm$ 1.23 & 87.62 $\pm$ 1.10 \\
    GCNII\textsuperscript{$\dagger$} & 88.49 $\pm$ 1.25  & 77.13 $\pm$ 1.48 & 90.30 $\pm$ 0.43  \\
    Geom-GCN\textsuperscript{$\dagger$} & 85.27 $\pm$ 1.57  & 77.99 $\pm$ 1.15  & 90.05 $\pm$ 0.47\\
    MixHop  & 87.61 $\pm$ 2.03 & 76.26 $\pm$ 2.95 & 85.31 $\pm$ 2.29 \\
    WRGAT & 88.20 $\pm$ 2.26 & 76.81 $\pm$ 1.89 & 88.52 $\pm$ 0.92 \\
    NSD\textsuperscript{$\dagger$}  & 87.14 $\pm$ 1.13  & 77.14 $\pm$ 1.57   & 89.49  $\pm$ 0.40\\
    GGCN  & 87.95 $\pm$ 1.05  & 77.14  $\pm$ 1.45  & 89.15 $\pm$ 0.37 \\
    H2GCN  & 87.87 $\pm$ 1.20  & 77.11 $\pm$ 1.57  & 89.49 $\pm$ 0.38  \\
    LINKX  & 84.64 $\pm$ 1.13 & 73.19 $\pm$ 0.99 & 87.86  $\pm$ 0.77 \\ 
    ACMII-GCN++ & 88.25 $\pm$ 0.96 & 77.12 $\pm$ 1.58 & 89.71 $\pm$ 0.48  \\
    \midrule
     ADR-GNN\textsubscript{S} &  {89.43 $\pm$ 1.15} &  {78.36 $\pm$ 144} & {90.55 $\pm$ 0.53}  \\
    \bottomrule
  \end{tabular}}
\end{table*}

\begin{table*}[t]
\centering
  \caption{Node classification accuracy ($ \%$) on \emph{heterophilic} datasets. $\dagger$ denotes the maximal accuracy of several proposed variants.}
  \label{table:heterophilic_fully_std}
  \begin{center}
     \footnotesize
   \setlength{\tabcolsep}{3pt}
  \begin{tabular}{ccccccc}
    \toprule
    Method & Squirrel & Film &  Cham. & Corn. & Texas & Wisc. \\
    Homophily & 0.22 & 0.22 & 0.23 & 0.30  & 0.11 & 0.21 \\
        \midrule
    GCN  & 23.96 $\pm$ 2.01 & 26.86  $\pm$  1.10 &  28.18  $\pm$  2.24&  52.70  $\pm$ 5.30  & 52.16  $\pm$ 5.16  & 48.92  $\pm$  3.06 \\
    GAT & 30.03  $\pm$  1.55 & 28.45  $\pm$ 0.89 & 42.93  $\pm$ 2.50 & 54.32  $\pm$ 5.05 & 58.38  $\pm$ 6.63 & 49.41  $\pm$ 4.09
    \\
    GCNII & 38.47  $\pm$ 1.58 & 32.87  $\pm$  1.30 &   60.61  $\pm$  3.04 & 74.86  $\pm$ 3.79 & 69.46  $\pm$ 3.83 & 74.12  $\pm$ 3.40 \\
    Geom-GCN\textsuperscript{$\dagger$}& 38.32  $\pm$ 0.92 & 31.63  $\pm$ 1.15 &  60.90  $\pm$ 2.81 & 60.81  $\pm$ 3.67 & 67.57  $\pm$ 2.72 & 64.12  $\pm$ 3.66 \\
    MixHop & 43.80  $\pm$  1.48 & 32.22  $\pm$ 2.34 & 60.50  $\pm$ 2.53 & 73.51  $\pm$ 6.34 & 77.84  $\pm$ 7.73 & 75.88  $\pm$ 4.90 \\ 
    GRAND & 40.05  $\pm$ 1.50 & 35.62  $\pm$ 1.01 &  54.67  $\pm$ 2.54 & 82.16  $\pm$ 7.09 & 75.68  $\pm$ 7.25 & 79.41  $\pm$ 3.64 \\ 
    NSD\textsuperscript{$\dagger$}  &  56.34  $\pm$  1.32 & {37.79  $\pm$ 1.15}  & 68.68  $\pm$ 1.58  &  86.49  $\pm$ 4.71 & 85.95  $\pm$ 5.51 & 89.41  $\pm$  4.74 \\
    WRGAT & 48.85  $\pm$ 0.78 & 36.53  $\pm$ 0.77 & 65.24  $\pm$ 0.87 &  81.62  $\pm$ 3.90 & 83.62  $\pm$ 5.50 & 86.98  $\pm$ 3.78 \\
    MagNet &  --  & -- & --  & 84.30  $\pm$ 7.00 & 83.30  $\pm$ 6.10 & 85.70  $\pm$ 3.20 \\ 
    GGCN  & 55.17  $\pm$ 1.58 & {37.81  $\pm$ 1.56} &  71.14  $\pm$ 1.84 & 85.68  $\pm$ 6.63  & 84.86  $\pm$ 4.55  &  86.86  $\pm$ 3.29 \\
    H2GCN  & 36.48  $\pm$ 1.86 & 35.70  $\pm$ 1.00 & 60.11  $\pm$ 1.71 & 82.70  $\pm$ 5.28  & 84.86  $\pm$  7.23 &  87.65  $\pm$ 4.98 \\
    GraphCON\textsuperscript{$\dagger$} & -- & -- & -- & 84.30  $\pm$  4.80 & 85.40  $\pm$ 4.20  & 87.80  $\pm$ 3.30 \\  
    FAGCN & 42.59  $\pm$ 0.69 & 34.87  $\pm$ 1.35 & 55.22  $\pm$ 2.11 & 79.19  $\pm$ 5.87 & 82.43  $\pm$ 2.87 & 82.94  $\pm$ 1.58 \\
    GPRGNN & 31.61  $\pm$ 1.24 & 34.63  $\pm$ 1.22 & 46.58  $\pm$ 1.71 & 80.27  $\pm$ 8.11 & 78.38  $\pm$ 4.36 & 82.94  $\pm$ 4.21 \\
    ACMP-GCN & --  & -- & -- & 85.40  $\pm$  7.00 & 86.20  $\pm$ 3.00 & 86.10  $\pm$ 4.00  \\ 
    LINKX  & 61.81  $\pm$ 1.80 &  36.10  $\pm$ 1.55 & 68.42  $\pm$ 1.38 & 77.84  $\pm$ 5.81 & 
    74.60   $\pm$  8.37 & 75.49  $\pm$ 5.72 \\
    GRAFF\textsuperscript{$\dagger$} & 59.01  $\pm$ 1.31 & 37.11  $\pm$ 1.08 & 71.38  $\pm$ 1.47 & 84.05  $\pm$ 6.10 & 88.38  $\pm$ 4.53 & 88.83  $\pm$ 3.29\\ 
    G\textsuperscript{2}\textsuperscript{$\dagger$} & {64.26  $\pm$ 2.38} &  37.30  $\pm$ 1.01 & {71.40  $\pm$ 2.38} & 87.30  $\pm$ 4.84 &  87.57  $\pm$ 3.86  & 87.84  $\pm$ 3.49\\
    ACMII-GCN++ & {67.40  $\pm$ 2.21} & 37.09   $\pm$ 1.32 & {74.76  $\pm$ 2.20} & 86.49  $\pm$ 6.73 & 88.38  $\pm$ 3.43 & 88.43  $\pm$ 3.66 \\
    \midrule
        ADR-GNN\textsubscript{S} & {72.54 $\pm$ 2.20} & {39.16 $\pm$ 1.13}  & {79.91 $\pm$ 2.27} & {91.89 $\pm$ 5.89} & {93.61 $\pm$ 4.26} & {93.46 $\pm$ 4.11} \\
    \bottomrule
  \end{tabular}
\end{center}
\end{table*}

\subsection{Large Scale Homophilic Node Classification}
We evaluate our ADR-GNN\textsubscript{S} on the OGBN-Arxiv dataset and compare it with several methods such as GCN, GRAND, FIE \citep{chen23fie}, GOAL \citep{zheng2023finding}, and ADGN \citep{zhao2021adaptive}, in Table \ref{table:arxiv}. The results further indicate the efficacy of ADR-GNN\textsubscript{S} on large scale homophilic node classification cases.

\begin{table*}[]
  \caption{Node classification accuracy ($ \%$) on OGBN-Arxiv.
  }   \label{table:arxiv}
  \begin{center}
     \footnotesize
  \begin{tabular}{lcccccc}
    \toprule
    Method & GCN & GRAND & FIE   & GOAL  & AGDN & ADR-GNN\textsubscript{S}  \\
    \midrule
    Accuracy (\%) & 71.74 & 72.23 & 72.39  & 71.25 & 73.41 & \textbf{73.68}  \\
\bottomrule
  \end{tabular}
\end{center}
\end{table*}

\subsection{Additional Heterophilic Node Classification Datasets}
\label{app:additional}
In addition to the 6 heterophilic node classification datasets reported in the main paper in Section \ref{sec:nodeExperiments}, we now also report the results on 4 additional heterophilic datasets from \cite{lim2021new, lim2021large}, to further demonstrate the effectiveness of our ADR-GNN\textsubscript{S}. The results are reported in Table \ref{tab:additionalHetero}, where we see that our ADR-GNN\textsubscript{S} achieves competitive results that are in line or better than other recent methods. In addition to the methods considered in the main paper, here we also consider SGC ([D] in the additional appendix references), L prop ([E]), and LINK \cite{lim2021new}.
\begin{table*}[ht]
    \centering
    \caption{Test accuracy on heterophilic datasets. For all datasets, we report the obtained accuracy (\%), besides Twitch-DE that considers test ROC AUC$\ddagger$. Standard deviations are over 5 train/val/test splits. Not available results are indicated by -- .}
    \label{tab:additionalHetero}
    {\footnotesize
    \center{
    \begin{tabular}{ccccc}
    \toprule
    Method & Twitch-DE$\ddagger$ &  Deezer-Europe & Penn94 (FB100)  & arXiv-year  \\
    \midrule
     MLP & 69.20 $\std$ 0.62 &  66.55 $\std$ {0.72} & 73.61 $\std$ 0.40 & 36.70 $\std$ 0.21   \\    
     \hline
     L Prop (2 hop) & 72.27 $\std$ 0.78   & 56.96 $\std$ 0.26 & 74.13 $\std$ 0.46 &   $46.07\std{0.15}$  \\
     LINK & 72.42 $\std$ {0.57} & 57.71 $\std$ 0.36 & 80.79 $\std$ {0.49}  &  53.97 $\std$ {0.18}  \\    
     LINKX & -- & -- & 84.41 $\std$ 0.52 & \third{56.00 $\std$ {1.34}} \\
     \hline
     SGC (2 hop) &  73.65 $\std$ {0.40} &  61.56 $\std$ {0.51} &  76.09 $\std$ 0.45 &   32.27 $\std$ {0.06}   \\    
     C\&S (2 hop) & 69.39 $\std$ 0.85 &  64.52 $\std$ 0.62 & 72.47 $\std$ 0.73 &    42.17 $\std$ 0.27 \\    
     \hline
     GCN &  \second{74.07 $\std$ 0.68}  & 62.23 $\std$ 0.53 &  82.47 $\std$ {0.27} &   46.02 $\std$ 0.26 \\
     GAT & 73.13 $\std$ 0.29 & 61.09 $\std$ {0.77} & 81.53 $\std$ {0.55}  &  46.05 $\std$ {0.51}  \\
     APPNP & 72.20 $\std$ {0.73} &  67.21 $\std$ {0.56} & 74.95 $\std$ {0.45} &   38.15 $\std$ {0.26}  \\
     \hline
     H2GCN & 72.67 $\std$ {0.65} &    67.22 $\std$ {0.90} & -- &    49.09 $\std$ {0.10}  \\
     GCNII & 72.38 $\std$ {0.31} & 66.42 $\std$ {0.56} & 82.92 $\std$ {0.59} & 47.21 $\std$ {0.28} \\
     MixHop & 73.23 $\std$ {0.99}   & 66.80 $\std$ {0.58} &  83.47 $\std${0.71} &      51.81 $\std$ {0.17}  \\
     GPR-GNN &  \third{73.84 $\std$ {0.69}}  &  \third{66.90 $\std$ {0.50}} &  \third{84.59 $\std$ {0.29}}   & 45.07 $\std$ {0.21}    \\
     ACMII\textsuperscript{$\dagger$} & -- & \second{67.50 $\std$ {0.53}} & \second{85.95 $\std$ {0.26}} & -- \\
     G\textsuperscript{2} & -- & -- & --  &\first{63.30$\pm$ 1.84}  \\
     \midrule
         ADR-GNN\textsubscript{S} & \first{74.98 $\pm$ 0.52} & \first{68.22 $\pm$ 0.57} & \first{86.63 $\pm$ 0.31} & \second{61.17 $\pm$ 1.54}
\\
    \bottomrule
    \end{tabular}}
    }
\end{table*}

\subsection{Spatio-Temporal Node Forecasting Results}
\label{app:spatioTemporal}
We provide here full results of our spatio-temporal experiments, both on the METR-LA and PEMS-BAY datasets. In addition, we also include results with 'classical' algorithms such as HA (historic averaging), VAR, and SVR. The results are reported in Table \ref{table:appmetrLA_pemsBay}.

\begin{table*}[t]
   \setlength{\tabcolsep}{3pt}
    \centering

    \label{tab:appTemporal}
    \begin{tabular}{ccccccccccc}

      \toprule
       \multirow{2}*{{Dataset}} &\multirow{2}*{{Method}} & \multicolumn{3}{c}{{Horizon 3}} & \multicolumn{3}{c}{{Horizon 6}}& \multicolumn{3}{c}{{Horizon 12}}\\ 
      \cmidrule(r){3-5} \cmidrule(r){6-8} \cmidrule(r){9-11}
      &  & MAE & RMSE & MAPE & MAE & RMSE & MAPE & MAE & RMSE & MAPE\\
      \midrule

      &HA              & 4.79  & 10.00 & 11.70\%       & 5.47  & 11.45 & 13.50\%      & 6.99  & 13.89  & 17.54\% \\ 
      &VAR             & 4.42  & 7.80  & 13.00\%       & 5.41  & 9.13  & 12.70\%      & 6.52  & 10.11 & 15.80\% \\ 
      &SVR             & 3.39  & 8.45  & 9.30\%        & 5.05  & 10.87 & 12.10\%      & 6.72  & 13.76 & 16.70\% \\ 
      &FC-LSTM         & 3.44  & 6.30  & 9.60\%        & 3.77  & 7.23  & 10.09\%      & 4.37  & 8.69  & 14.00\% \\ 
      &DCRNN           & 2.77  & 5.38  & 7.30\%        & 3.15  & 6.45  & 8.80\%       & 3.60  & 7.60  & 10.50\% \\ 
      &STGCN           & 2.88  & 5.74  & 7.62\%        & 3.47  & 7.24  & 9.57\%       & 4.59  & 9.40  & 12.70\% \\ 
      {METR}&Graph WaveNet   & 2.69  & \third{5.15}  & 6.90\%        & 3.07  & 6.22  & 8.37\%       & 3.53  & 7.37  & 10.01\% \\
      {-LA}&ASTGCN          & 4.86  & 9.27  & 9.21\%        & 5.43  & 10.61 & 10.13\%      & 6.51  & 12.52 & 11.64\% \\  
      &STSGCN          & 3.31  & 7.62  & 8.06\%        & 4.13  & 9.77  & 10.29\%      & 5.06  & 11.66 & 12.91\% \\  
      &GMAN            & 2.80  & 5.55  & 7.41\%        & 3.12  & 6.49  & 8.73\%       & \third{3.44}  & 7.35  & 10.07\% \\  
      &MTGNN           & 2.69  & 5.18  & \third{6.88\%}        & 3.05  & \third{6.17}  & \third{8.19\%}       & 3.49  & \third{7.23}  & \third{9.87\%} \\  
      &GTS             & \third{2.67}  & 5.27  & 7.21\%        & \third{3.04}  & 6.25  & 8.41\%       & 3.46  & 7.31  & 9.98\% \\  
    &STEP      & \second{2.61}  & \second{4.98} & \second{6.60\%}        & \second{2.96}  & \second{5.97}  & \second{7.96\%}      & \second{3.37}  & \second{6.99}  & \second{9.61\%} \\ 
        & STAEformer   & 2.65 & 5.11 & 6.85\% & 2.97 & 6.00 & 8.13\% & 3.34 & 7.02 & 9.70\% \\
       \cmidrule(r){2-11}
    &   ADR-GNN\textsubscript{T} & \first{2.53} & \first{4.85} & \first{6.51\%} &  \first{2.81}  & \first{5.82} & \first{7.39\%}  & \first{3.19} & \first{6.89}  & \first{9.10\%} \\
    \midrule
      &HA              & 1.89  & 4.30  & 4.16\%        & 2.50  & 5.82  & 5.62\%       & 3.31  & 7.54  & 7.65\% \\ 
      &VAR             & 1.74  & 3.16  & 3.60\%        & 2.32  & 4.25  & 5.00\%       & 2.93  & 5.44  & 6.50\% \\ 
      &SVR             & 1.85  & 3.59  & 3.80\%        & 2.48  & 5.18  & 5.50\%       & 3.28  & 7.08  & 8.00\% \\ 
      &FC-LSTM         & 2.05  & 4.19  & 4.80\%        & 2.20  & 4.55  & 5.20\%       & 2.37  & 4.96  & 5.70\% \\ 
      &DCRNN           & 1.38  & 2.95  & 2.90\%        & 1.74  & 3.97  & 3.90\%       & 2.07  & 4.74  & 4.90\% \\ 
      &STGCN           & 1.36  & 2.96  & 2.90\%        & 1.81  & 4.27  & 4.17\%       & 2.49  & 5.69  & 5.79\% \\ 
      {PEMS}&Graph WaveNet   & \third{1.30}  & \third{2.74}  & \third{2.73\%}        & \third{1.63}  & \third{3.70}  & \third{3.67\%}       & 1.95  & 4.52  & 4.63\% \\
      {-BAY}&ASTGCN          & 1.52  & 3.13  & 3.22\%        & 2.01  & 4.27  & 4.48\%       & 2.61  & 5.42  & 6.00\% \\  
      &STSGCN          & 1.44  & 3.01  & 3.04\%        & 1.83  & 4.18  & 4.17\%       & 2.26  & 5.21  & 5.40\% \\  
      &GMAN            & 1.34  & 2.91  & 2.86\%        & \third{1.63}  & 3.76  & 3.68\%       & \third{1.86}  & \third{4.32}  & \third{4.37\%} \\  
      &GTS             & 1.34  & 2.83  & 2.82\%        & 1.66  & 3.78  & 3.77\%       & 1.95  & 4.43  & 4.58\% \\  
      &STEP      & \second{1.26}  & \second{2.73}  & \second{2.59\%}        & \second{1.55}  & \second{3.58}  & \second{3.43\%}     & \second{1.79}  & \second{4.20}  & \second{4.18\%} \\ 
          & STAEformer   &1.31 & 2.78 & 2.76\% & 1.62 & 3.68 & 3.62\% & 1.88 & 4.34 & 4.41\%
\\
    \cmidrule(r){2-11}
    &  ADR-GNN\textsubscript{T} & \first{1.13}  & \first{2.36} & \first{2.30\%} & \first{1.39} & \first{3.13} & \first{3.01\%} & \first{1.68} & \first{3.81} & \first{3.82\%}
 \\
      \bottomrule
    \end{tabular}
        \caption{Multivariate time series forecasting on the METR-LA, and PEMS-BAY datasets.} \label{table:appmetrLA_pemsBay}
  \end{table*}

\subsection{Synthetic Experiment}
\label{app:synthetic}
We now provide the details of the construction of the synthetic experiment from our ablation study in Section \ref{sec:ablation} in the main paper. The experiment is conducted as follows: we generate a random Erd\H{o}s-R\'{e}nyi graph $\mathcal{G}_{\rm{ER}}=(\mathcal{E}_{\rm{ER}}, \mathcal{V}_{\rm{ER}})$, and randomly select a set of \emph{source} nodes $\mathcal{V}_{\rm{ER}}^{\rm{src}} \subset \mathcal{V}_{\rm{ER}}$  assigned with value of $\frac{1}{|\mathcal{V}_{\rm{ER}}^{\rm{src}}|}$. The rest of the nodes $\mathcal{V}_{\rm{ER}} \setminus \mathcal{V}_{\rm{ER}}^{\rm{src}}$ are initialized with a value of 0. We also choose a random node $\bfv_{\rm{dst}} \in \mathcal{V}_{\rm{ER}} \setminus \mathcal{V}_{\rm{ER}}^{\rm{src}}$. The goal is to transport all the  mass from all the nodes to the $\bfv_{\rm{dst}}$, such that $\bfv_{\rm{dst}}$ will have a feature of 1, the the rest of the nodes in the graph will be zeroes. That is, all the features in the graph are concentrated in $\bfv_{\rm{dst}}$. In our example in Figure \ref{fig:syntheticExample}, we use the protocol specified here, to generate a graph with 5 nodes, and we show the approximation obtained with the advection, diffusion, and reaction terms.

\subsection{The Dirichlet Energy of ADR-GNN}
\label{app:oversmoothing}
We follow \cite{rusch2022gradient} and define the Dirichlet energy of the graph node features as:
\begin{equation}
    \label{eq:gatEnergy}
    {\rm{E}}(\bfU^{(l)}) = \frac{1}{|\mathcal{V}|}\sum_{i \in \mathcal{V}} \sum_{j \in \mathcal{N}_i} ||\bfU^{(l)}_i - \bfU^{(l)}_j ||_2^2. 
\end{equation}

Following the experiment in our ablation study in Section \ref{sec:ablation}, we now also report the measured Dirichlet energy of our ADR-GNN\textsubscript{S} on Cora and Citeseer, with 64 layers. We also compare the measured Dirichlet energy of GCN, for reference. The results are reported in Figure \ref{fig:dirichletEnergy}, where we show the relative (to the initial node features $\bfU^{(0)}$) Dirichlet energy. It is evident that ADR-GNN\textsubscript{S} does not oversmooth, because the energy does not decay to 0, as in GCN.

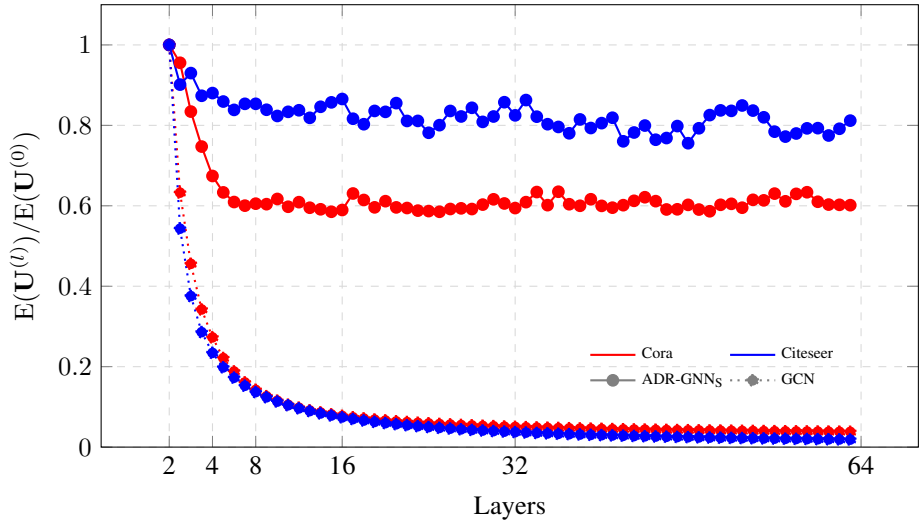
\begin{figure*}[t]
    \centering
  \begin{minipage}[t]{0.7\textwidth}
    \centering
\begin{tikzpicture}
  \begin{axis}[
      width=1.0\linewidth, 
      height=0.6\linewidth,
      grid=major,
      grid style={dashed,gray!30},
      xlabel=Layers,
      ylabel=${\rm E}(\bfU^{(l)}) / {\rm E}(\bfU^{(0)})$,
      ylabel near ticks,
      legend style={at={(0.75,0.25)},anchor=north,scale=0.8, draw=none, cells={anchor=west}, font=\tiny, fill=none},
      legend columns=2,
     xtick={0,4,8,16,32,64},
      xticklabels = {2,4,8,16,32,64},
          yticklabel style={
            /pgf/number format/fixed,
            /pgf/number format/precision=3
          },
      scaled y ticks=false,
      every axis plot post/.style={thick},
      ymin = 0
    ]
    \addplot[red, mark=oplus*, forget plot]
    table[x=layer
,y=adr_cora
,col sep=comma] {data/adr_energies.csv};
    \addplot[blue, mark=oplus*, forget plot]
    table[x=layer
,y=adr_citeseer
,col sep=comma] {data/adr_energies.csv};
    \addplot[red, style=dotted, mark=oplus*, forget plot]
    table[x=layer
,y=gcn_cora
,col sep=comma] {data/adr_energies.csv};
    \addplot[blue, style=dotted, mark=oplus*, forget plot]
    table[x=layer
,y=gcn_citeseer
,col sep=comma] {data/adr_energies.csv};
    \addplot[red, draw=none] coordinates {(1,1)};
    \addplot[blue, draw=none] coordinates {(1,1)};
    \addplot[gray, draw=none,mark=oplus*] coordinates {(-10,-10)};
    \addplot[gray, style=dotted, draw=none,mark=oplus*] coordinates {(-10,-10)};
    \legend{Cora, Citeseer, ADR-GNN\textsubscript{S}, GCN}
    \end{axis}
\end{tikzpicture}
\captionof{figure}{The relative Dirichlet energy  vs. model depth. 
}
\label{fig:dirichletEnergy}
  \end{minipage}
\end{figure*}

\cleardoublepage

\section*{Additional Appendix References}

[A] Kaiming He, Xiangyu Zhang, Shaoqing Ren, and Jian Sun. Deep residual learning for
 image recognition. In Proceedings of the IEEE Conference on Computer Vision and Pattern Recognition, pages 770–778, 2016.

[B] R. A. Horn and C. R. Johnson. Matrix Analysis. Academic Press, 1985.

[C] Ashish Vaswani, Noam Shazeer, Niki Parmar, Jakob Uszkoreit, Llion Jones, Aidan N Gomez,
 Łukasz Kaiser, and Illia Polosukhin. Attention is all you need. Advances in neural information
 processing systems, 30, 2017.

 [D] Felix Wu, Amauri Souza, Tianyi Zhang, Christopher Fifty, Tao Yu, and Kilian Weinberger.
 Simplifying graph convolutional networks. In International conference on machine learning,
 pages 6861–6871. PMLR, 2019.

 [E] Leto Peel. Graph-based semi-supervised learning for relational networks. In Proceedings of
910 the 2017 SIAM international conference on data mining, pages 435–443. SIAM, 2017.

\end{document}